%% file: main.tex
\documentclass[10pt,twocolumn,letterpaper]{article}

\usepackage{cvpr}              %
\usepackage{amssymb}%
\usepackage{pifont}%
\newcommand{\greencheck}{\textcolor{green}{\ding{51}}}
\newcommand{\redcheck}{\textcolor{red}{\ding{55}}}
\usepackage{multirow}
\usepackage{multicol}
\usepackage{colortbl}
\usepackage{footnote}
\usepackage{float} 

\input{preamble}

\definecolor{cvprblue}{rgb}{0.21,0.49,0.74}
\usepackage[pagebackref,breaklinks,colorlinks,citecolor=cvprblue]{hyperref}

\title{\name: A Large-Scale Scene Dataset for Deep Learning-based 3D Vision}

\author{Lu Ling$^1$, \quad Yichen Sheng$^{1}$, \quad Zhi Tu$^{1}$, \quad Wentian Zhao$^{2}$, \quad Cheng Xin$^{3}$, \quad Kun Wan $^{2}$, \\
\quad Lantao Yu$^{2}$, \quad Qianyu Guo$^{1}$, \quad Zixun Yu $^{4}$, \quad Yawen Lu$^{1}$, \quad Xuanmao Li$^{5}$, \\
\quad Xingpeng Sun $^{1}$, \quad Rohan Ashok$^{1}$, \quad Aniruddha Mukherjee$^{1}$, \quad Hao Kang $^{6}$, \quad Xiangrui Kong$^{1}$,\\
 Gang Hua$^{6}$, \quad Tianyi Zhang$^{1}$, \quad Bedrich Benes$^{1}$ , \quad Aniket Bera$^{1}$\\
\textsuperscript{1} \small Department of Computer Science, Purdue University,
\quad \textsuperscript{2} \small Adobe Inc. ,
\quad \textsuperscript{3} \small Rutgers University \\
\textsuperscript{4} \small Google Inc. ,
\quad \textsuperscript{5} \small Huazhong University of Science and Technology,
\quad \textsuperscript{6} \small Wormpex AI Research\\
}

\input{sec/macros}

\begin{document}
\twocolumn[{%
\renewcommand\twocolumn[1][]{#1}%
\maketitle
\vspace{-5mm}
\input{sec/0_teaser}

\bigbreak
}]

\input{sec/0_abstract}    
\input{sec/1_Introduction}

\input{sec/2_Related_work}

\input{sec/3_Data_Statistic}

\input{sec/4_Benchmark}

\input{sec/5_Experiment}
\input{sec/6_Conclusion}

{
    \small
    \bibliographystyle{ieeenat_fullname}
    \bibliography{main}
}

\input{sec/X_suppl}

\end{document}

%% file: preamble.tex
\usepackage[dvipsnames]{xcolor}

%% file: sec/macros.tex
\usepackage{color}
\usepackage{soul}
\usepackage{multirow}
\usepackage{xcolor}

\newlength\savewidth

\definecolor{turquoise}{cmyk}{0.65,0,0.1,0.1}
\definecolor{purple}{rgb}{0.65,0,0.65}
\definecolor{darkgreen}{rgb}{0.0, 0.5, 0.0}
\definecolor{darkred}{rgb}{0.5, 0.0, 0.0}
\definecolor{darkblue}{rgb}{0.0, 0.0, 0.5}
\definecolor{blue}{rgb}{0.0, 0.0, 1.0}
\definecolor{orange}{rgb}{1.0,0.5,0.0}

\newcommand{\hide}[1]{{}}

\makeatletter
\renewcommand{\paragraph}{%
  \@startsection{paragraph}{4}%
  {\z@}{0.3ex \@plus 1ex \@minus .1ex}{-1em}%
  {\normalfont\normalsize\bfseries}%
}
\makeatother

\newif\ifproofread

\definecolor{tabfirst}{rgb}{1, 0.7, 0.7} 
\definecolor{tabsecond}{rgb}{1, 0.85, 0.7}
\definecolor{tabthird}{rgb}{1, 1, 0.7}

\newcommand{\name}{{\textit{DL3DV-10K}}}
\newcommand{\benchmark}{{\textit{DL3DV-140}}}
\newcommand{\scale}{{\textbf{10,510}}}

%% file: sec/0_teaser.tex
\setlength{\tabcolsep}{0pt}
\begin{tabular}{c}
\includegraphics[width=\linewidth]{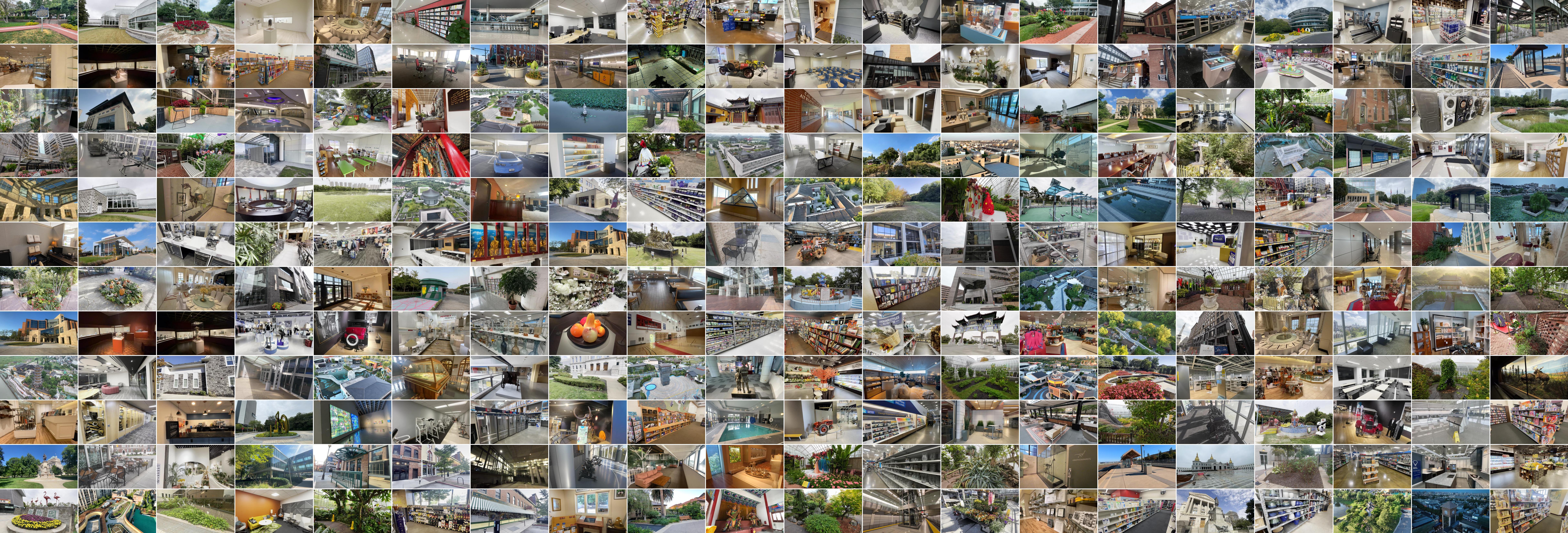} \\
\end{tabular}
\captionof{figure}{We introduce \textbf{{\name}}, a large-scale, scene dataset capturing real-world scenarios. {\name} contains {\scale} videos at 4K resolution spanning 65 types of point-of-interest (POI) locations, covering a wide range of everyday areas. With the fine-grained annotation on scene diversity and complexity, {\name} enables a comprehensive benchmark for novel view synthesis and supports learning-based 3D representation techniques in acquiring a universal prior at scale.}
\label{fig:teaser}

%% file: sec/0_abstract.tex
\begin{abstract}
We have witnessed significant progress in deep learning-based 3D vision, ranging from neural radiance field (NeRF) based 3D representation learning to applications in novel view synthesis (NVS). However, existing scene-level datasets for deep learning-based 3D vision, limited to either synthetic environments or a narrow selection of real-world scenes, are quite insufficient. This insufficiency not only hinders a comprehensive benchmark of existing methods but also caps what could be explored in deep learning-based 3D analysis. To address this critical gap, we present \name, a large-scale scene dataset, featuring \textbf{51.2 million} frames from \scale\ videos captured from \textbf{65} types of point-of-interest (POI) locations, covering both bounded and unbounded scenes, with different levels of reflection, transparency, and lighting. We conducted a comprehensive benchmark of recent NVS methods on \name, which revealed valuable insights for future research in NVS. In addition, we have obtained encouraging results in a pilot study to learn generalizable NeRF from \name, which manifests the necessity of a large-scale scene-level dataset to forge a path toward a foundation model for learning 3D representation. Our \name\ dataset, benchmark results, and models will be publicly accessible at \href{https://dl3dv-10k.github.io/DL3DV-10K/}{Project Page}.

\end{abstract}

%% file: sec/1_Introduction.tex
\section{Introduction}

The evolution in deep 3D representation learning, driven by essential datasets, boosts various tasks in 3D vision. 
Notably, the inception of Neural Radiance Fields~\cite{mildenhall2021nerf} (NeRF), offering a new approach through a continuous high-dimensional neural network, revolutionized leaning-based 3D representation and novel view synthesis (NVS). NeRF excels at producing detailed and realistic views, overcoming challenges faced by traditional 3D reconstruction methods and rendering techniques. Additionally, it inspires waves of innovative developments such as NeRF variants~\cite{barron2022mip,barron2023zip,tancik2023nerfstudio,xu2022point,chen2021mvsnerf,yu2021plenoctrees} and 3D Gaussian Splatting (3DGS)~\cite{kerbl20233d}, significantly enhancing experiences in virtual reality, augmented reality, and advanced simulations.

However, existing scene-level datasets for NVS are restricted to either synthetic environments or a narrow selection of real-world scenes due to the laborious work for scene collection. Notably, the absence of such large-scale scene datasets hinders the potential of deep 3D representation learning methods in two pivotal aspects: 1)~it is impossible to conduct a comprehensive benchmark to adequately assess NVS methods in complex real-world scenarios such as non-Lambertian surfaces. 2)~It restricts the generalizability of deep 3D representation learning methods on attaining universal priors from substantial real scenes.

To fill this gap, we revisited the commonly used dataset for benchmarking NVS. i)~Synthetic datasets like blender dataset~\cite{mildenhall2021nerf} offer rich 3D CAD geometry but lack real-world elements, which diminishes the model's robustness in practical applications. ii)~Real-world scene datasets for NVS such as Tank and temples~\cite{knapitsch2017tanks} and LLFF dataset~\cite{mildenhall2019local} offer more variety but limited scope. They fail to capture the complex real-world scenarios such as intricate lighting and various material properties (transparency and reflectance~\cite{rematas2016deep}), which still challenges current SOTAs. 

Moreover, existing 3D representation methods yield photorealistic views by independently optimizing NeRFs for each scene, requiring numerous calibrated views and substantial compute time. Learning-based models like PixlNeRF~\cite{yu2021pixelnerf} and IBRNet~\cite{wang2021ibrnet} mitigate this by training across multiple scenes to learn scene priors.  While datasets like RealEstate~\cite{zhou2018stereo} and ScanNet++~\cite{yeshwanth2023scannet++} improve understanding in specific domains such as indoor layouts, their limited scope hinders the broader applicability of these models. This is primarily due to the absence of comprehensive scene-level datasets, which are crucial for learning-based methods to achieve universal representation.

Based on the above review, 
We introduce \name\, a novel dataset that captures large-scale (multi-view) MV scenes using standard commercial cameras to enable efficient collection of a substantial variety of real-world scenarios. \name\ comprises \textbf{51.3} million frames from \scale\ videos in 4k resolution, covers scenes from 65 types of point-of-interest~\cite{ye2011exploiting} (POI) locations like restaurants, tourist spots, shopping malls, and natural outdoor areas. Each scene is further annotated with its complexity indices, including indoor or outdoor environments, the level of reflection and transparency, lighting conditions, and the level of texture frequency. Fig.~\ref{fig:teaser} provides a glimpse of our \name\ dataset. Tab.~\ref{tab:dataset_comparison} compares scale, quality, diversity, and annotated complexity between \name and the existing scene-level datasets.

Additionally, we present \benchmark, a comprehensive benchmark for NVS, by sampling 140 scenes from the dataset. The diversity and fine-grained scene complexity in \benchmark\ will enable a fair evaluation of NVS methods. We conducted the statistical evaluation of the SOTA NVS methods on \benchmark\ (Sec.~\ref{subsec:NVS_experiment}), including Nerfacto~\cite{tancik2023nerfstudio}, Instant-NGP~\cite{muller2022instant}, Mip-NeRF 360~\cite{barron2022mip}, Zip-NeRF~\cite{barron2023zip}, and 3DGS~\cite{kerbl20233d}. 
Leveraging the multi-view nature of the data, we attempt to showcase \name's potential for deep 3D representation learning methods in gaining a universal prior for generating novel views.
Our demonstrations indicate that pretraining on \name\  enhances the generalizability of NeRF  (Sec.~\ref{subsec:Gene_NeRF}), confirming that diversity and scale are crucial for learning a universal scene prior. 

To summarize, we present the following contributions:

\begin{enumerate}
    \item We introduce \name, a real-world MV scene dataset. It has 4K resolution RGB images and covers 65 POI indoor and outdoor locations. Each scene is annotated with the POI category, light condition, environment setting, varying reflection and transparency, and the level of texture frequency. 
    \item We provide \benchmark, a comprehensive benchmark with 140 scenes covering the challenging real-world scenarios for NVS methods. We conduct the statistical evaluation for the SOTA NVS methods on \benchmark\ and compare their weaknesses and strengths. 
    \item We show that the pre-training on \name\ benefits generalizable NeRF to attain universal scene prior and shared knowledge. 
\end{enumerate}

%% file: sec/2_Related_work.tex
\section{Related Work}\label{sec:realted_work}

\begin{table*}[t]
\centering
\footnotesize
\begin{tabular}{lrccr|ccc|c}
\toprule
\multirow{2}{*}{Dataset} & \multirow{2}{*}{\# of scene} & \multirow{2}{*}{\# of POI category} & \multirow{2}{*}{Resolution} & \multicolumn{1}{c|}{\multirow{2}{*}{\# of frames}} & \multicolumn{3}{c|}{Scene complexity annotation}    & \multicolumn{1}{c}{\multirow{2}{*}{NVS}} \\
                         &                        &                      &                             & \multicolumn{1}{c|}{}                     & indoor/outdoor & reflection & transparency & \multicolumn{1}{c}{}                     \\ \midrule

                         LLFF~\cite{mildenhall2019local}& 24 & - &640 $\times$ 480 & $<$1K & \greencheck \ \greencheck &\greencheck&\redcheck&\greencheck \\ \hline
                         DTU~\cite{jensen2014large} & 124  &5 &1200 $\times$ 1600&30K&\greencheck \ \redcheck&\greencheck&\redcheck&\greencheck \\ \hline
                        BlendedMVS~\cite{yao2020blendedmvs} & 113 &-&1536 $\times$ 2048&17K&\greencheck \ \greencheck&\redcheck&\redcheck& \redcheck \\ \hline
                        ScanNet~\cite{dai2017scannet} &1513&11&1296 $\times$ 968&2,500K&\greencheck \ \redcheck&\redcheck&\redcheck&\redcheck \\ \hline
                        Matterport3D~\cite{chang2017matterport3d} &90\footnotemark[1] &-&1280 $\times$ 1024&195K&\greencheck \   \redcheck&\greencheck&\redcheck&\redcheck \\ \hline
                        Tanks and Temples~\cite{knapitsch2017tanks} &21&14&3840 $\times$ 2160&147K&\greencheck \   \greencheck&\greencheck&\redcheck&\greencheck \\ \hline
                        ETH3D~\cite{schops2017multi}&25&11&6048 $\times$ 4032&$<$1K&\greencheck \   \greencheck&\redcheck&\redcheck&\redcheck \\ \hline
                        RealEstate10K\footnotemark[2]~\cite{zhou2018stereo} & 10,000 &1&1280 $\times$ 720& 10,000K& \greencheck \ \redcheck & \redcheck & \redcheck & \greencheck\\ \hline
                        ARKitScenes~\cite{baruch2021arkitscenes}& 1661&1&1920 $\times$ 1440&450K&\greencheck \   \redcheck&\redcheck&\redcheck&\redcheck \\ \hline
                            ScanNet++~\cite{yeshwanth2023scannet++} &460&5& 7008 $\times$ 4672\footnotemark[3] &3,980K&\greencheck \   \redcheck &\greencheck&\redcheck&\greencheck \\ \hline
                        \name \ (ours)& \scale & 65 & 3840 $\times$ 2160 & 51,200K & \greencheck \   \greencheck &\greencheck&\greencheck& \greencheck \\ 
        \bottomrule      
\end{tabular}
\caption{Comparison of the existing scene-level dataset in terms of quantity, quality, diversity, and complexity, which is measured by the fine-grained surface properties, light conditions, texture frequency, and environmental setting. 
}
\label{tab:dataset_comparison}
\end{table*}

\subsection{Novel View Synthesis}
\textbf{Novel View Synthesis Methods.} 
Early novel view synthesis (NVS) work concentrated on 3D geometry and image-based rendering~\cite{levoy2023light,zhou2018stereo}. Since 2020, Neural Radiance Fields (NeRF)~\cite{mildenhall2021nerf} have been pivotal in NVS for their intricate scene representation, converting 5D coordinates to color and density, leading to various advancements in rendering speed and producing photorealistic views. Key developments include Instant-NGP~\cite{muller2022instant}, which speeds up NeRF using hash tables and multi-resolution grids; Mip-NeRF~\cite{barron2021mip} addressed aliasing and Mip-NeRF 360~\cite{barron2022mip} expanded this to unbounded scenes with significant computational power; and Zip-NeRF~\cite{barron2023zip} combines Mip-NeRF 360 with grid-based models for improving efficiency. Nerfacto~\cite{tancik2023nerfstudio} merges elements from previous NeRF methods to balance speed and quality. Additionally, 3D Gaussian Splatting (3DGS)~\cite{kerbl20233d} uses Gaussian functions for real-time, high-quality rendering. 

However, those SOTA methods, building neural radiance fields for individual scenes, require dense views and extensive computation. Learning-based models like ContraNeRF~\cite{yang2023contranerf}, TransNeRF~\cite{wang2022generalizable}, PixelNeRF~\cite{yu2021pixelnerf}, and IBRNet~\cite{wang2021ibrnet} overcome this by training on numerous scenes for universal priors and sparse-view synthesis. Yet, the absence of large-scale scene-level datasets reflective of real-world diversity fails to provide adequate assessment for the SOTAs. Additionally, it hinders the potential for learning-based 3D models to gain a universal prior.

\paragraph{Novel View Synthesis Benchmarks.} 
NVS benchmarks are generally split into synthetic benchmarks like the NeRF-synthetic (Blender)~\cite{mildenhall2021nerf}, ShapeNet~\cite{chang2015shapenet} and Objaverse~\cite{deitke2023objaverse}, featuring 3D CAD models with varied textures and complex geometries but lacking real-world hints such as noise and non-Lambertian effects.

In contrast, real-world NVS benchmarks, originally introduced for multi-view stereo (MVS) tasks like DTU~\cite{jensen2014large} and Tanks and Temples~\cite{knapitsch2017tanks}, offer limited variety. While ScanNet~\cite{dai2017scannet} has been used for benchmarking NVS, its motion blur and narrow field-of-view limit its effectiveness. Later benchmarks for outward- and forward-facing scenes emerged, but they vary in collection standards and lack diversity, particularly in challenging scenarios like lighting effects on reflective surfaces. For example, LLFF~\cite{mildenhall2019local} offers cellphone-captured 24 forward-facing scenes; Mip-NeRF 360 dataset~\cite{mildenhall2021nerf} provides 9 indoor and outdoor scenes with uniform distance around central subjects; Nerfstudio dataset~\cite{tancik2023nerfstudio} proposes 10 captures from both phone and mirrorless cameras with different lenses.

Inspired by the capture styles of these datasets, \benchmark\ provides a variety of scenes to comprehensively evaluate NVS methods, including challenging view-dependent effects, reflective and transparent materials, outdoor (unbounded) environments, and high-frequency textures. We also offer extensive comparative analyses, demonstrating the efficacy of \benchmark\ in assessing NVS techniques.

\subsection{Multi-view Scene Dataset} 
\footnotetext[1]{90 building-scale scenes covering 2056 rooms}
\footnotetext[2]{YouTube video}
\footnotetext[3]{7008$\times$4672 in 270 scenes and  1920$\times$1440 in 190 scenes}

Multi-view (MV) datasets are commonly used for NVS tasks in the 3D vision field. These datasets range from synthetic, like ShapeNet~\cite{chang2015shapenet} and Pix2Vox++~\cite{xie2020pix2vox++}, to foundational real-world datasets capturing both object-level and scene-level images. 

Object-level datasets like Objectron~\cite{ahmadyan2021objectron}, CO3D~\cite{reizenstein2021common}, and MVimgnet~\cite{yu2023mvimgnet} offer substantial scale for learning-based reconstruction. While they facilitate the learning of spatial regularities, reliance solely on object-level MV datasets impedes prediction performance on unseen objects and complex scenes containing multiple objects~\cite{yu2021pixelnerf}. 

Scene-level datasets are essential for NVS and scene understanding, yet offerings like LLFF~\cite{mildenhall2019local}, Matterport3D~\cite{chang2017matterport3d}, and BlendedMVS~\cite{yao2020blendedmvs} encompass limited scenes. DTU~\cite{jensen2014large}, despite its use in developing generalized NeRF~\cite{yu2021pixelnerf}, is limited by its scale, affecting the models' ability to generalize. RealEstate10k~\cite{zhou2018stereo}, ARKitScenes~\cite{baruch2021arkitscenes}, ScanNet~\cite{dai2017scannet}, and the high-resolution ScanNet++~\cite{yeshwanth2023scannet++} improve this with a broad range of detailed indoor scenes. Yet, their applicability remains less comprehensive for diverse indoor settings like shopping centers and restaurants, or outdoor scenarios. Although RealEstate10k, focusing on YouTube real estate videos, provides comparable scale with us, they comprise low resolution and lack of diversity. Overall, the limited scale and diversity of these datasets pose challenges for the robust and universal training of 3D deep learning models. We close this gap by introducing \name, encompassing multifarious real-world scenes from indoor to outdoor environments, enhancing 3D spatial perception, and paving the way for more robust learning-based 3D models.

%% file: sec/3_Data_Statistic.tex
\begin{figure}[t]
    \centering
    \includegraphics[width=\linewidth]{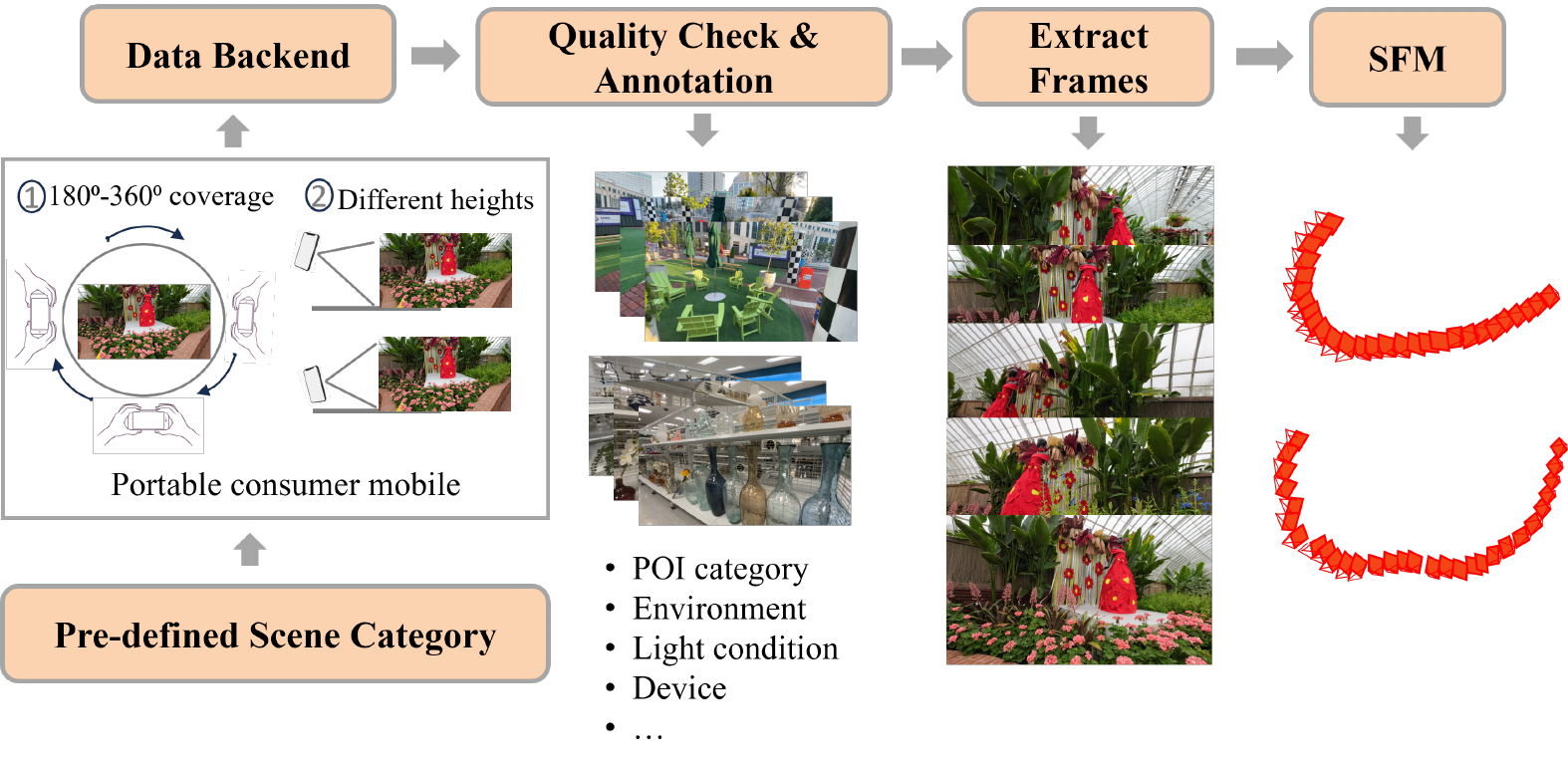}
    \caption{The efficient data acquisition pipeline of \name. Refer to \textit{supplementary materials} for more visual illustrations of scene coverage.}
    \label{fig:pipeline}
\end{figure}

\section{Data Acquisition and Processing}

Our data acquisition goal is to gather large-scale, high-quality scenes reflective of real-world complexity and diversity. We develop a pipeline that integrates video capture, pre-processing, and analysis, leveraging widely available consumer mobiles and drones to ensure the coverage of everyday accessible areas. We designed a detailed guideline to train the collectors to minimize motion blur, exclude exposure lights, and avoid moving objects. This collection process, illustrated in Fig.~\ref{fig:pipeline}, is user-friendly and enhances both quality and efficiency.

\begin{figure*}[t]
    \centering
    \includegraphics[width=1.02\textwidth]{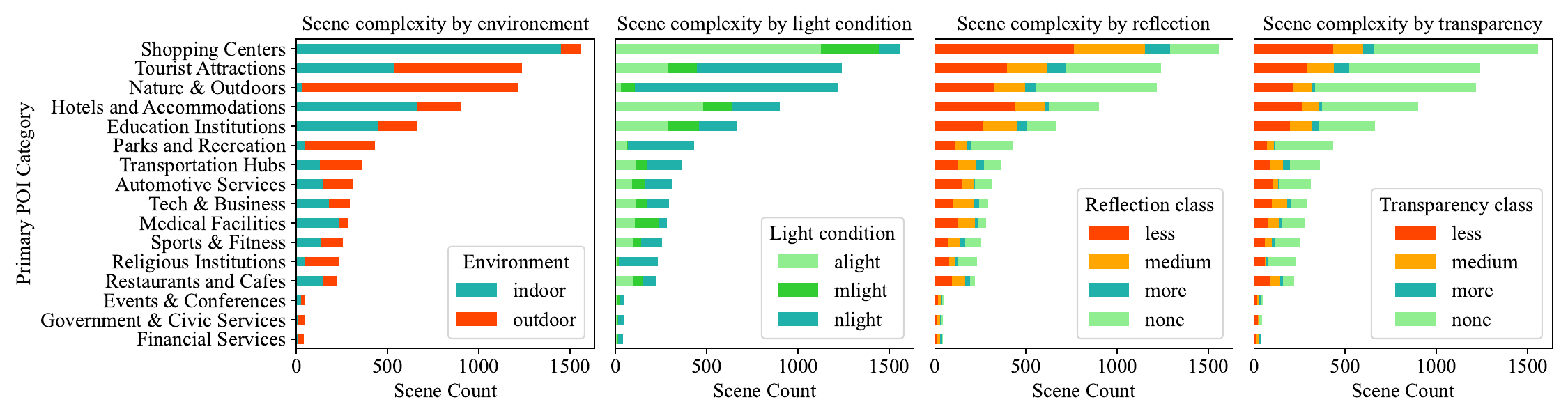}
    \caption{ We show the distribution of scene category (the primary POI locations) by complexity indices, including environmental setting, light conditions, reflective surface, and transparent materials. Attributes in light conditions include: natural light (`\textit{nlight}'), artificial light (`\textit{alight}'), and a combination of both (`\textit{mlight}'). Reflection class includes `\textit{more}', `\textit{medium}', `\textit{less}', and `\textit{none}'. Transparency class likewise.
    }
    \label{fig:scene_complexity_by_POI}
\end{figure*}

\subsection{Data Acquisition}
\label{subsec:data_acq}

\paragraph{Diversity.} 
Our collection focuses on various commonly accessible scenes, adhering to the point-of-interest (POI)~\cite{ye2011exploiting} categories. \name\ captures scenes from 16 primary and 65 secondary POI categories. These categories include varied settings such as educational institutions, tourist attractions, restaurants, medical facilities, transportation hubs, etc. The diversity, spanning from indoor to outdoor environments and different cities, is instrumental in enriching the dataset with a broader spectrum of cultural and architectural variations.
Such variety is essential for benchmarking NVS techniques as it challenges and refines their ability to generalize across a multitude of real-world scenarios and fosters robustness and adaptability in 3D models. 
Sec.~\ref{sec:stat} summarizes the POI category of collected scenes.

\paragraph{Complexity.} Real-world scenes challenge state-of-the-art techniques (SOTAs) with their diverse environments (indoor vs. outdoor), view-dependent lighting effects, reflective surfaces, material properties, and high-frequency textures. The high detail frequency in textures, from delicate fabric to coarse stone, requires meticulous rendering. Outdoor (unbounded) scenes, with varying details between near and distant objects, challenge the robustness of NVS methods in handling scale differences. Complex shadows and view-dependent highlights from natural and artificial lights, interacting with reflective and transparent materials like metal and glass, require precise handling for realistic depiction. 
Additionally, we provide multiple views of the scene from various angles at different heights and distances. This multiplicity and complexity of views enable 3D methods to predict view-dependent surface properties. It also aids in separating viewpoint effects in learning view-invariant representations like patch descriptors~\cite{zeng20173dmatch} and normals~\cite{zhang2017physically}.

\paragraph{Quality.}
The quality of the video is measured by the content and coverage density of the scene and video resolution. We have formulated the following requirements as guidelines for recording high-quality scene-level videos: 1)~The scene coverage is in the circle or half-circle with a 30 secs-45 secs walking diameter and has at least five instances with a natural arrangement. 2)~The default focal length of the camera corresponds to the 0.5x ultra-wide mode for capturing a wide range of background information. 3)~Each video should encompass a horizontal view of at least 180$^{\circ}$ or 360$^{\circ}$ from different heights, including overhead and waist levels. It offers high-density views of objects within the coverage area. 4)~The video resolution should be 4K and have 60 fps (or 30 fps). 5)~The video's length should be at least 60 secs for mobile phone capture and 45 secs for drone video recording. 6)~We recommend limiting the duration of moving objects in the video to under 3 secs, with a maximum allowance of 10 secs. 7)~The frames should not be motion-blurred or overexposed. 8)~The captured objects should be stereoscopic. Post-capture, we meticulously inspect videos based on the above criteria, such as moving objects in the video, to guarantee the high quality of videos captured on mobile devices. Tab.~\ref{tab:device} summarizes the number of scenes recorded by different devices and the associated quality levels in terms of moving objects' duration in the videos.

\begin{table}[t]
\footnotesize
\begin{tabular}{c|cc|cc}
\hline
\multirow{2}{*}{Category} & \multicolumn{2}{c|}{Device}             & \multicolumn{2}{c}{Quality by moving objects}                                   \\ \cline{2-5} 
                           & \multicolumn{1}{c|}{Consumer mobile} & Drone & \multicolumn{1}{c|}{\textless{}3s } & {3s - 10s } \\ \hline
\# of scene            & \multicolumn{1}{c|}{10,407}        & 103   & \multicolumn{1}{c|}{8064}                        & 2446                 \\ \hline
\end{tabular}
\caption{Number of scenes by devices and level of quality.}
\label{tab:device}
\end{table}

\subsection{Data Processing}

\paragraph{Frequency Estimation.}
To estimate the frequency metric of the scene over the duration of the captured video, we first sample $100$ frames from each video, as texture frequency is typically calculated based on RGB images. Then, we convert RGB images to grayscale and 
normalize the intensities.
To extract the high-frequency energy, we apply a two-dimensional bi-orthogonal wavelet transform~\cite{cohen1992biorthogonal} and compute the Frobenius norm of the ensemble of LH, HL and HH subbands. The Frobenius norm is finally normalized by the number of pixels, and the average quotient over $100$ frames is the frequency metric. The distribution of frequency metrics is shown in \textit{supplementary materials}.

\paragraph{Labeling.}
We categorize and annotate the diversity and complexity of scenes based on our established criteria, including key attributes of the scene, such as POI category, device model, lighting conditions, environmental setting, surface characteristics, and high-frequent textures. The POI category and device model depict the location where the video was collected and the equipment used during its capture. Lighting conditions are differentiated into natural, artificial, or a combination of both, influencing the ambient illumination of scenes. The environmental setting annotation distinguishes between indoor and outdoor settings, which is crucial for evaluating the NVS performance in both bounded and unbounded spaces. We measure the surface properties by the level of reflectivity, ranging from more and medium to less and none. It is estimated by the ratio of reflective pixels in the image and its present duration in the video. Material property, measured by transparency, follows a similar rule.
Refer to \textit{supplementary materials} for more details on reflection and transparency labeling criteria.

\subsection
{
Data Statistics}\label{sec:stat}
\paragraph{Scale.} \name\ aims to provide comprehensive and diverse coverage of scene-level datasets for 3D vision. It covers scenes collected from 16 primary POIs and 65 secondary POIs and comprises 51.3 million frames of \scale\ videos with 4K resolution. As shown in Tab.~\ref{tab:dataset_comparison}, it enjoys fine-grained annotation for scene complexity indices.

\paragraph{Hierarchy.}
We classify the scene category following the POI category taxonomy. For example, the primary POI category is the `\textit{entertainment area}'. Its secondary POI category includes `\textit{theaters}', `\textit{concert halls}', `\textit{sports stadiums}', and `\textit{parks and recreation areas}'. The statistics of primary POI-based scene categories by annotated complexity are presented in Fig.~\ref{fig:scene_complexity_by_POI}. The distribution of scenes captured in these POI locations follows: 1) their generality in nature. 2) 
the probability of no moving objects appearing within 60 sec in the locations. 3) the accessibility to these locations. For example, the government and civic services locations usually do not allow video shooting in high-level details. 4) the collected video is under the permission of the agents.

%% file: sec/4_Benchmark.tex
\subsection{Benchmark}\label{sec:benchmark}
To comprehensively assess the SOTAs, the benchmark needs to cover the inherent complexity of real-world scenarios with varying reflectance, texture, and geometric properties. To achieve our goal, we present \benchmark\ as a comprehensive benchmark, sampling 140 static scenes from our dataset. Additionally, we simplified the reflection and transparency classes into two categories for better interpretation: more reflection (including more and medium reflection) and less reflection (including less and no reflection); the same approach applies to transparency. \benchmark\, with scenes collected from diverse POIs, maintains a balance in each annotated scene complexity indices. This means \benchmark\ are categorized as \textit{indoor} (bounded) scenes vs. \textit{outdoor} (unbounded) scenes, high vs. low texture frequency (\textit{low-freq vs. high-freq}), more vs. less reflection (\textit{more-ref vs. less-ref}), and more vs. less transparency (\textit{more-transp vs. less-transp}). 
\benchmark\ offers challenging scenes with a rich mix of diversity and complexity for a comprehensive evaluation of existing SOTA methods.

%% file: sec/5_Experiment.tex
\begin{figure*}[t]
    \centering
    \includegraphics[width=\textwidth]{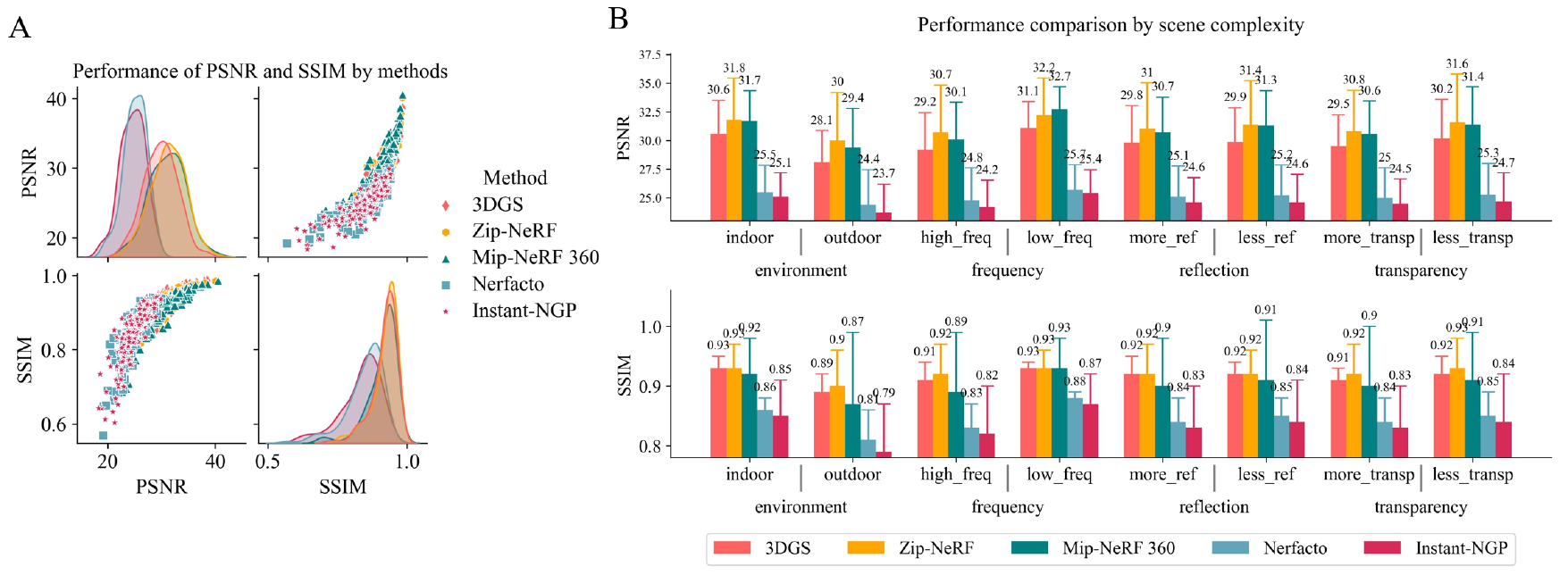}
    \caption{A presents the density plot of PSNR and SSIM and their relationship on \benchmark~for each method. B describes the performance comparison by scene complexity. The text above the bar plot is the mean value of the methods on the attribute.  }
    \label{fig:overall_Metric}
\end{figure*}

\begin{table}[t]
    \centering
    \footnotesize
    \begin{tabular}{l|cccccc}
    & PSNR $\uparrow$ & SSIM $\uparrow$ & LPIPS $\downarrow$ & Train  $\downarrow$ &  Mem  $\downarrow$ \\ 
    \hline
Instant-NGP  &                      25.01 &                      0.834 &                      0.228 &  \cellcolor{tabfirst}1.2 hr& \cellcolor{tabsecond}3.9GB \\
Nerfacto     &                      24.61 &                      0.848 &                      0.211 &                      2.6 hr&  \cellcolor{tabfirst}3.7GB \\
Mip-NeRF 360 & \cellcolor{tabsecond}30.98 &  \cellcolor{tabthird}0.911 &  \cellcolor{tabthird}0.132 &                      48.0 hr&                      23.6GB \\
3DGS         &  \cellcolor{tabthird}29.82 & \cellcolor{tabsecond}0.919 & \cellcolor{tabsecond}0.120 & \cellcolor{tabsecond}2.1 hr&  \cellcolor{tabthird}16.8GB \\
Zip-NeRF*    &                      29.07 &                      0.878 &                      0.169 &  \cellcolor{tabthird}2.5 hr&                      23.8GB \\
Zip-NeRF     &  \cellcolor{tabfirst}31.22 &  \cellcolor{tabfirst}0.921 &  \cellcolor{tabfirst}0.112 &                      4.0 hr&                      38.2GB \\
    \end{tabular}
    \caption{Performance on \benchmark. The error metric is calculated from the mean of 140 scenes on a scale factor of 4. Zip-NeRF uses the default batch size (65536) and Zip-NeRF* uses the identical batch size as other methods (4096). Note, the training time and memory usage may be different depending on various configurations. Refer to \textit{supplementary materials} for details.}
    \label{tab:average_metric}
\end{table}

\section{Experiment} \label{sec:Experiment}
\subsection{Evaluation on the NVS benchmark} \label{subsec:NVS_experiment}

\paragraph{Methods for comparison.}
We examine the current relevant state-of-the-art (SOTA) NVS methods on \benchmark, including NeRF variants such as Nerfacto~\cite{tancik2023nerfstudio}, Instant-NGP~\cite{muller2022instant}, Mip-NeRF 360~\cite{barron2022mip}, Zip-NeRF~\cite{barron2023zip}, and 3D Gaussian Splatting (3DGS)~\cite{kerbl20233d}.

\paragraph{Experiment details.}
The SOTAs have different assumptions on the image resolution.
For fairness, we use 960$\times$560 resolution to train and evaluate all the methods. 
Each scene in the benchmark has 300-380 images, depending on the scene size. 
We use 7/8 of the images for training and 1/8 of the images for testing. 
Different methods vary in their method configurations. 
We use most of the default settings, like network layers and size, optimizers, etc. 
But to ensure fairness, we fix some standard configurations. 
Each NeRF-based method (Nerfacto, Instant-NGP, Mip-NeRF 360, Zip-NeRF) has the same 4,096 ray samples per batch (equivalent to chunk size or batch size), the same near 0.05 and far 1$e$6. 
Note that Mip-NeRF 360 and Zip-NeRF use a much higher number of rays (65,536) per batch by default.  
We modify the learning rate to match the change of ray samples as suggested by the authors.
We notice that Zip-NeRF performance is sensitive to the ray samples.
So, we add one more experiment for Zip-NeRF with the same ray samples of 4,096 as other methods.
For all methods, we train enough iterations until they converge.

\paragraph{Quantitative results.}
Tab.~\ref{tab:average_metric} summarizes the average PSNR, SSIM, and L-PIPS metrics across all scenes in \benchmark\, along with training hours and memory consumption for each method. 
Furthermore, Fig.~\ref{fig:overall_Metric} provides detailed insights into the metric density functions and their correlations. 

The results indicate that Zip-NeRF, Mip-NeRF 360, and 3DGS consistently outperform Instant-NGP and Nerfacto across all evaluation metrics. 
Remarkably, Zip-NeRF demonstrates superior performance in terms of average PSNR and SSIM, although it consumes more GPU memory using the default batch size. Besides, we notice that reducing the default batch size for Zip-NeRF significantly decreases its PSNR, SSIM, and LPIPS, see Zip-NeRF* in  Tab.~\ref{tab:average_metric}. 
Mip-NeRF 360 achieves a PSNR of 30.98 and SSIM of 0.91, yet it shows relatively lower computational efficiency, with an average training time of 48 hours.
The density functions of PSNR and SSIM, depicted in 
Fig.~\ref{fig:overall_Metric}A, underscores Zip-NeRF and 3DGS's robust performance across all scenes. Moreover, we observe that 3DGS, with an SSIM of 0.92, surpasses Mip-NeRF 360's SSIM of 0.91, consistent with the findings from 3DGS's evaluation using the Mip-NeRF 360 dataset~\cite{kerbl20233d}.

Fig.~\ref{fig:overall_Metric} B illustrates the performance across scene complexity indices. Among all indices, outdoor (unbounded) scenes appear to be the most challenging, as all methods yield the lowest PSNR and SSIM scores in this setting. Conversely, low-frequency scenes are the easiest to generate.  Furthermore, more transparent scenes present higher challenges compared to less transparent ones. In terms of method comparison, Zip-NeRF outperforms others in most scenes, except in low-frequency scenarios where Mip-NeRF 360 demonstrates superior performance. Additionally, Mip-NeRF 360's smaller standard deviation in low-frequency scenes indicates its robustness in this scenario. We also present the findings of SOTAs' performance in terms of scene diversity, which are described in the \textit{supplementary materials}.

\begin{figure*}[t]
    \centering
    \includegraphics[width=\linewidth]{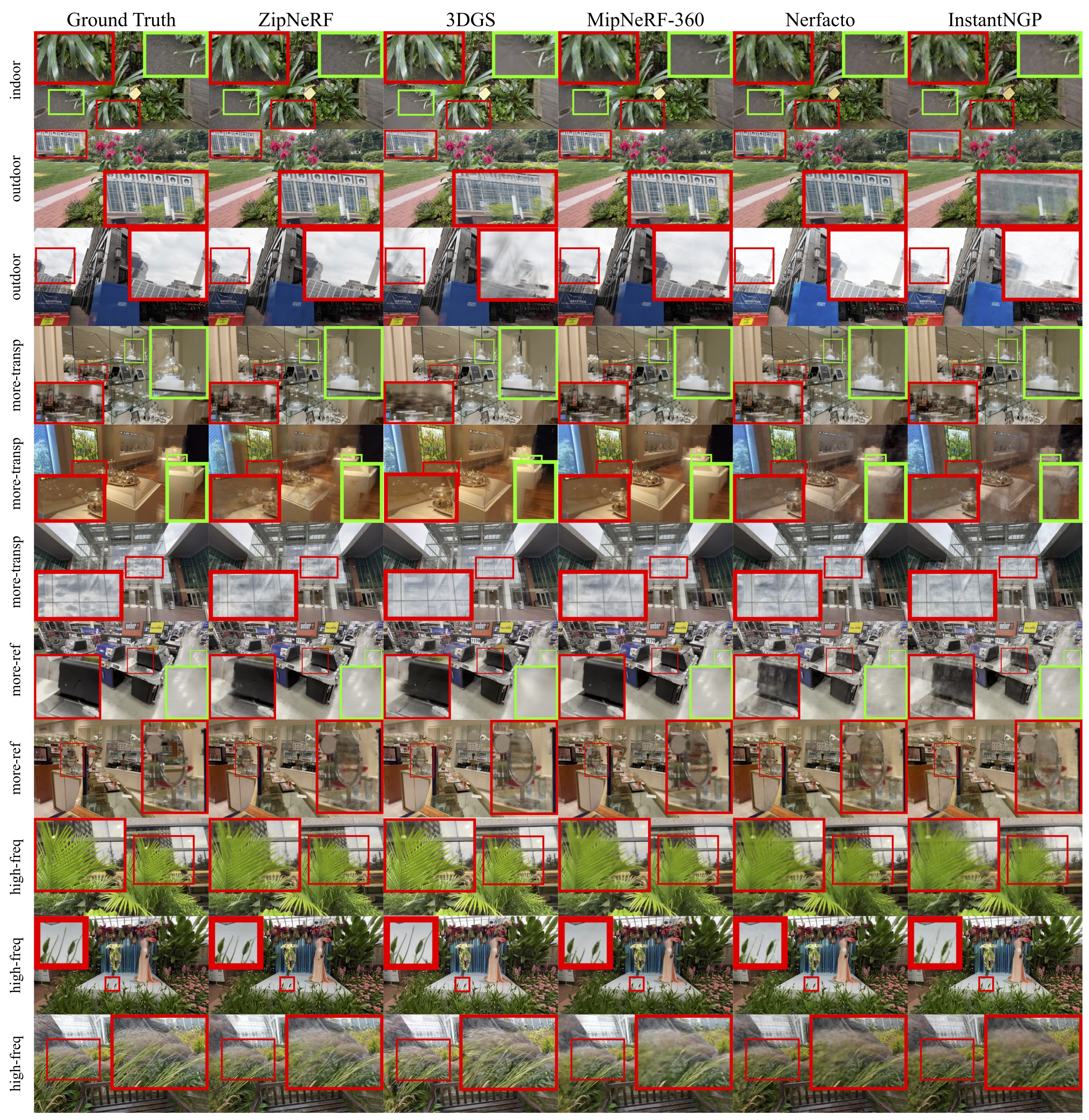}
    \caption{We compare the SOTA NVS methods and the corresponding ground truth images on \benchmark\ from held-out test views. More examples can be found in \textit{supplementary materials}. The scenes are classified by complexity indices: \textit{indoor vs. outdoor}, \textit{more-ref vs. less-ref}, \textit{high-freq vs. low-freq}, and \textit{more-transp vs. less-transp}. Best view by zooming in.}
    \label{fig:qualitative_comparison}
\end{figure*}

\paragraph{Visual results.} We show the visual result for the SOTAs on \benchmark\ in Fig.~\ref{fig:qualitative_comparison}.  
Overall, the artifact pattern for NeRF variants is the amount of ``grainy” microstructure, while 3DGS creates elongated artifacts or ``splotchy” Gaussians.

NeRF variants exhibit high sensitivity to distance scale, often generating blurry backgrounds and less-detailed foregrounds. For instance, Instant-NGP produces floating artifacts in the far-distance background of unbounded scenes. Although Zip-NeRF and Mip-NeRF 360 output fine-grained details compared to other NeRF variants, they also struggle with aliasing issues in sharp objects with high-frequent details, such as grasses and tiny leaves, as shown in Fig.~\ref{fig:qualitative_comparison} with `\textit{high-freq}' case. In contrast, 3DGS performs better against aliasing issues than NeRF variants; it suffers from noticeable artifacts in the far-distance background, such as the sky and far-distance buildings, as shown in Fig.~\ref{fig:qualitative_comparison} with `\textit{outdoor}' case. 

Regarding view-dependent effects in reflective and transparent scenes, 3DGS excels at rendering finely detailed and sharp lighting, such as strong reflections on metal or glass surfaces, and effectively captures subtle edges of transparent objects, a challenge for other methods. However, it tends to oversimplify softer reflective effects, like cloud reflections on windows or subtle light on the ground, as shown in Fig.~\ref{fig:qualitative_comparison} with `\textit{more-ref}' and `\textit{more-transp}' cases. In contrast, Zip-NeRF and Mip-NeRF 360 are less sensitive to the intensity of reflective light, capturing reflections more generally. On the other hand, Nerfacto and Instant-NGP struggle with these complex lighting effects, often producing floating artifacts.

\subsection{Generalizable NeRF}
\label{subsec:Gene_NeRF}
Recent NeRFs and 3DGS aim to only fit the training scene. 
Using these NVS methods to generalize to unseen real-world scenarios requires training on a large set of real-world multi-view images.
Due to the lack of real-world scene-level multi-view images, existing works either resort to training on large-scale object-level synthetic data~\cite{chen2021mvsnerf,trevithick2021grf,yu2021pixelnerf,rematas2021sharf,chibane2021stereo} or a hybrid of synthetic data and a small amount of real data~\cite{yang2023contranerf,wang2022generalizable,wang2021ibrnet}. 
The limited real-world data cannot fully bridge the domain gap. 
In this section, we conduct a pilot experiment to show that our large-scale real-world scene \name\ dataset has the potential to drive the learning-based generalizable NeRF methods by providing substantial real-world scenes for training.    

\paragraph{Experiment details.}
We choose IBRNet~\cite{wang2021ibrnet} as the baseline to conduct an empirical study.  
To demonstrate the effectiveness of \name, we pre-train the IBRNet on our \name\ to obtain a general prior and fine-tune on the training dataset used by IBRNet and compare the performance with the train-from-scratch IBRNet on the evaluation datasets used by IBRNet. 
ScanNet++~\cite{yeshwanth2023scannet++} is another recent high-quality real-world scene dataset that focuses on indoor scenarios. 
We conduct a similar experiment on ScanNet++ to further show that the richer diversity and larger scale of \name\ significantly improve the generalizable NeRFs results.  

\begin{figure}[H]
    \centering
    \includegraphics[width=\linewidth]{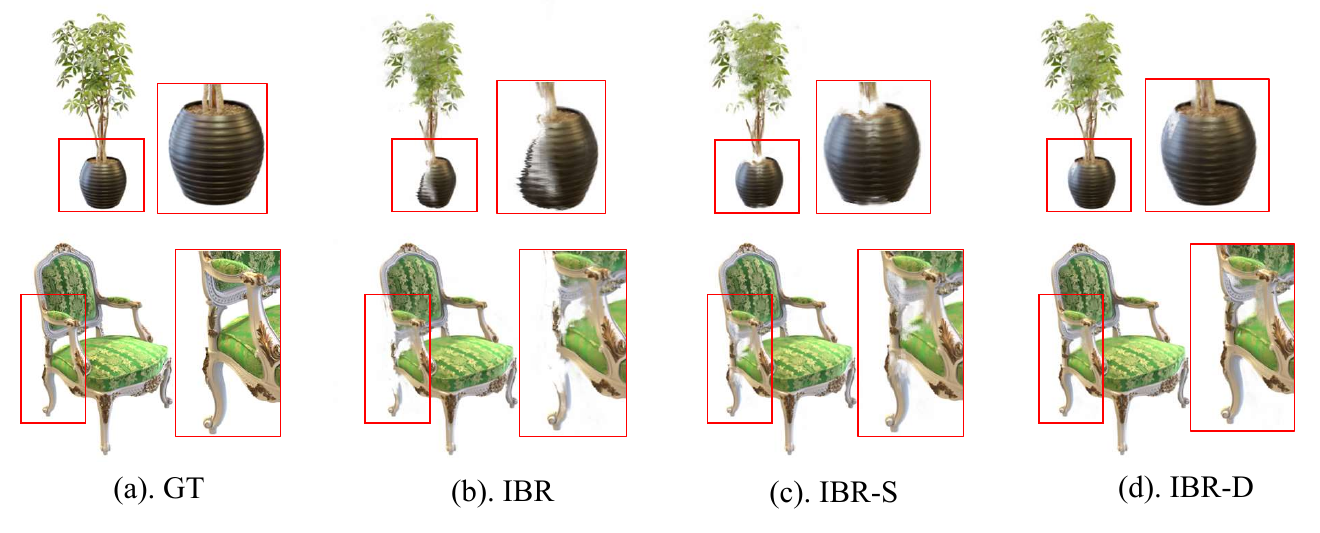}
    \caption{Qualitative comparison of (a). Ground truth. (b). IBRNet trained from scratch. (c). IBRNet pretrained on ScanNet++ and (d). IBRNet pretrained on DL3DV-2K}
    \label{fig:ibr_qualitative}
\end{figure}

\newcommand{\tablespace}{\,\,\,\,}
\newcommand{\halftablespace}{\,}
\setlength{\tabcolsep}{4pt}
\begin{table}[t]
\centering
\tiny %
\begin{tabular}{l|ccc|ccc|ccc}
\toprule
 & \multicolumn{3}{c|}{Diffuse Synthetic $360^\circ$~\cite{hedman2018deep}} & \multicolumn{3}{c|}{Realistic Synthetic $360^\circ$~\cite{mildenhall2021nerf}} & \multicolumn{3}{c}{Real Forward-Facing~\cite{mildenhall2019local}} \\
Method & PSNR$\uparrow$ & SSIM$\uparrow$ & LPIPS$\downarrow$ & PSNR$\uparrow$ & SSIM$\uparrow$ & LPIPS$\downarrow$ & PSNR$\uparrow$ & SSIM$\uparrow$ & LPIPS$\downarrow$ \\
\midrule
IBRNet     &                      34.72 & \cellcolor{tabsecond}0.983 & \cellcolor{tabsecond}0.024 &  \cellcolor{tabthird}23.95 &  \cellcolor{tabthird}0.906 &  \cellcolor{tabthird}0.101 &                      24.82 &                      0.808 &  \cellcolor{tabthird}0.178 \\
IBRNet-S   &                      34.22 &  \cellcolor{tabthird}0.979 & \cellcolor{tabsecond}0.024 &                      23.57 &                      0.905 &  \cellcolor{tabthird}0.101 &                      24.86 &                      0.807 &                      0.183 \\
IBRNet-270 & \cellcolor{tabsecond}35.18 &  \cellcolor{tabfirst}0.984 & \cellcolor{tabsecond}0.024 & \cellcolor{tabsecond}24.55 & \cellcolor{tabsecond}0.911 & \cellcolor{tabsecond}0.097 &  \cellcolor{tabthird}25.00 &  \cellcolor{tabthird}0.812 &                      0.180 \\
IBRNet-1K  &  \cellcolor{tabthird}35.13 &  \cellcolor{tabfirst}0.984 &  \cellcolor{tabfirst}0.023 &                      23.58 &  \cellcolor{tabthird}0.906 &                      0.102 & \cellcolor{tabsecond}25.02 & \cellcolor{tabsecond}0.814 &  \cellcolor{tabfirst}0.175 \\
IBRNet-2K  &  \cellcolor{tabfirst}35.34 &  \cellcolor{tabfirst}0.984 & \cellcolor{tabsecond}0.024 &  \cellcolor{tabfirst}24.98 &  \cellcolor{tabfirst}0.913 &  \cellcolor{tabfirst}0.095 &  \cellcolor{tabfirst}25.08 &  \cellcolor{tabfirst}0.815 & \cellcolor{tabsecond}0.176 \\
\bottomrule
\end{tabular}
\caption{IBRNet: IBRNet trained from scratch, IBRNet-S: IBRNet trained from ScanNet++, IBRNet-270, IBRNet-1K, and IBRNet-2K: IBRNet trained from 270 scenes, 1,000 scenes, and 2,000 scenes from \name. Refer to \textit{supplementary materials} for more samples on \name.}
\label{table:ibr_results}  
\end{table}
\setlength{\tabcolsep}{1.4pt}

\paragraph{Results.}
The quantitative and qualitative results are shown in Tab.~\ref{table:ibr_results} and Fig.~\ref{fig:ibr_qualitative}.   
The knowledge learned from ScanNet++ does not help IBRNet perform better on these existing benchmarks.       
However, the prior learned from a subset of our \name\ help IBRNet perform better on all the evaluation benchmarks.  
Also, when an increased data input from \name\, IBRNet consistently performs better in all the benchmarks. 

%% file: sec/6_Conclusion.tex
\section{Conclusion}
We introduce \name, a large-scale multi-view scene dataset, gathered by capturing high-resolution videos of real-world scenarios. The abundant diversity and the fine-grained scene complexity within \benchmark, featuring 140 scenes from \name, create a challenging benchmark for Neural View Synthesis (NVS). Our thorough statistical evaluation of SOTA NVS methods on \benchmark\ provides a comprehensive understanding of the strengths and weaknesses of these techniques. Furthermore, we demonstrate that leveraging \name\ enhances the generalizability of NeRF, enabling the development of a universal prior. This underscores the potential of \name\ in paving the way for the creation of a foundational model for learning 3D representations.

\paragraph{Limitations.}
\name\ encompasses extensive real-world scenes, enjoying the coverage of everyday accessible areas. This rich diversity and scale provide valuable insights for exploring deep 3D representation learning. However, there are certain limitations. While we demonstrate \name's potential in static view synthesis, some scenes include moving objects due to the nature of mobile phone video scene collection, as classified in Tab.~\ref{tab:device}, thereby introducing additional challenges for NVS. Nonetheless, such challenges may provide insights into exploring the robustness of learning-based 3D models. Moreover, these challenges may be solved by future learning-based 3D models for dynamic NVS.

%% file: sec/X_suppl.tex
\clearpage
\setcounter{page}{1}
\maketitlesupplementary

\section{Overview}
Our supplementary materials include this \textbf{pdf}, \textbf{video demo} and a \textbf{HTML} for more qualitative results.

Sec.~\ref{sec.data} discusses the data acquisition standard and the \name\ data distribution.  
Sec.~\ref{sec.exp} discusses more details of our benchmark experiments, including experiment, training details and more qualitative results, and details of the generalizable NeRF experiment.

\section{Data} \label{sec.data}
\subsection{Data acquisition}

\begin{figure*}[h]
    \centering
    \includegraphics[width=0.8\textwidth]{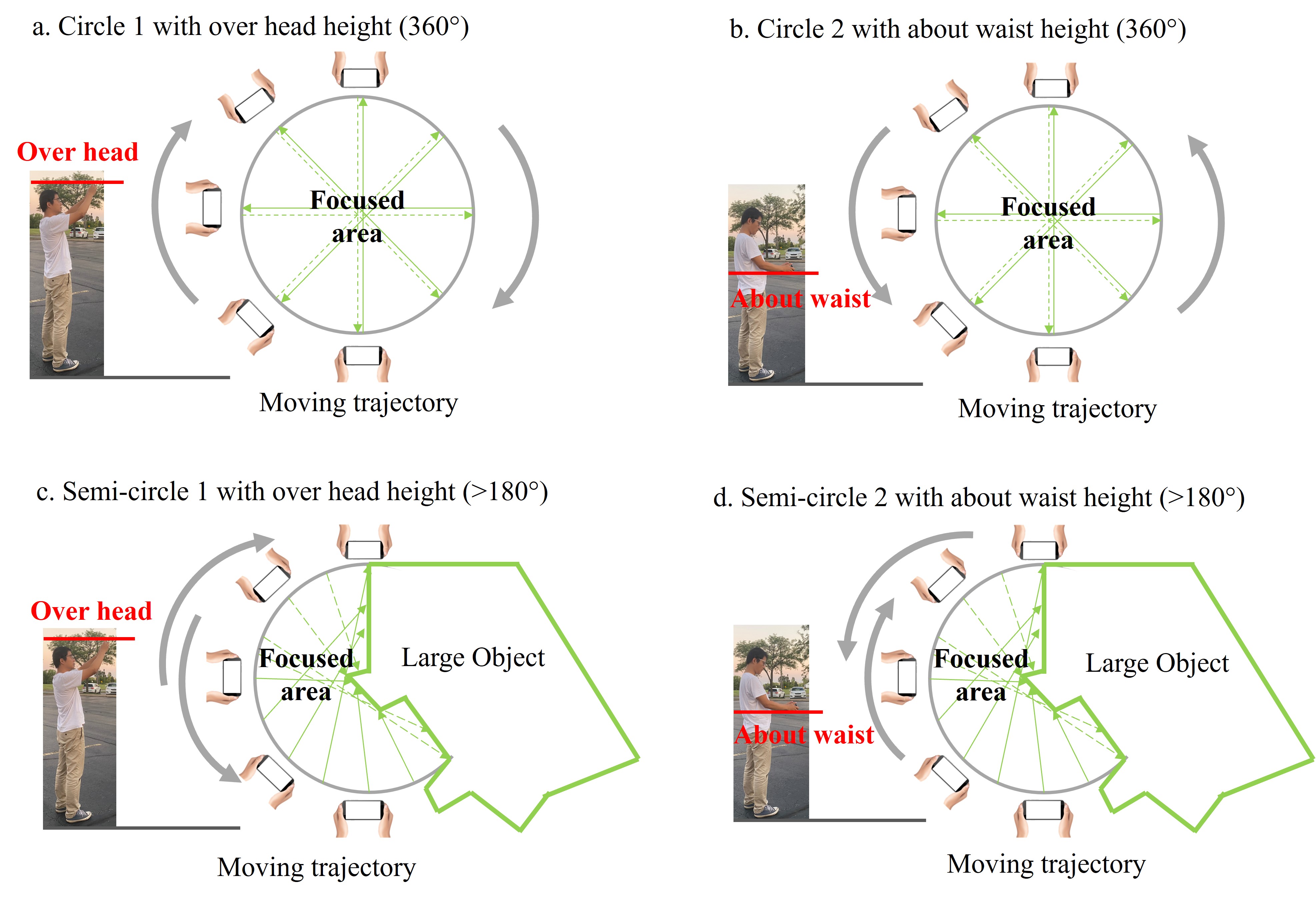}
    \caption{Video shooting examples with different heights and angles.}
    \label{fig:video_shooting_inst}
\end{figure*}

The scene coverage for video shooting is illustrated in Fig.~\ref{fig:video_shooting_inst}. For real-world scenes, they encompass horizontal views (180$^{\circ}$ - 360$^{\circ}$) from different heights. 
We capture scenes using 360$^{\circ}$ panoramic views when the scene is accessible and well-defined, typically encompassing a diameter that can be covered on foot within 30 to 45 secs. In instances where the rear view of the scene is obstructed by larger objects, such as larger buildings, we opt for a semi-circular view (exceeding 180$^{\circ}$) to capture the scene. To enhance scene coverage, we record videos by traversing two circular or semi-circular paths. The first traversal is conducted at overhead height, while the second is performed at approximately waist height.

\subsection{Labeling}

\begin{figure}[h]
    \centering
    \includegraphics[width=\linewidth]{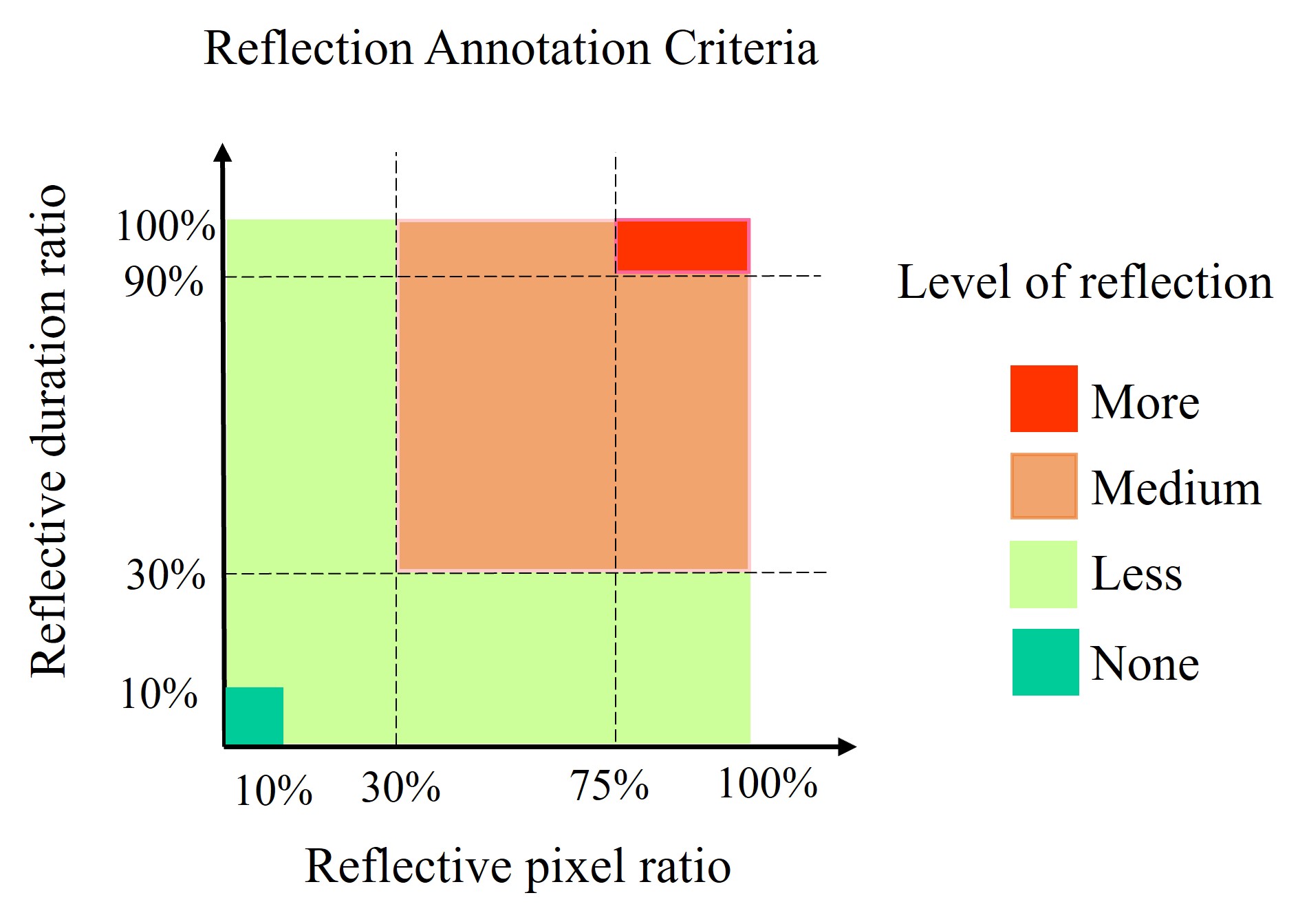}
    \caption{Reflection labeling criteria. Transparency annotation likewise.}
    \label{fig:reflection-label}
\end{figure}

\paragraph{Reflection and Transparency}
We manually annotate reflection and transparency indices to scenes by assessing the ratio of reflective (transparent) pixels and the duration of reflectivity (transparency) observed in the video. Fig.~\ref{fig:reflection-label} presents the reflection labeling criteria. Transparency labeling follows the same rule.

\subsection{Data Statistics}
\begin{figure*}[t]
    \centering
    \includegraphics[width=\linewidth]{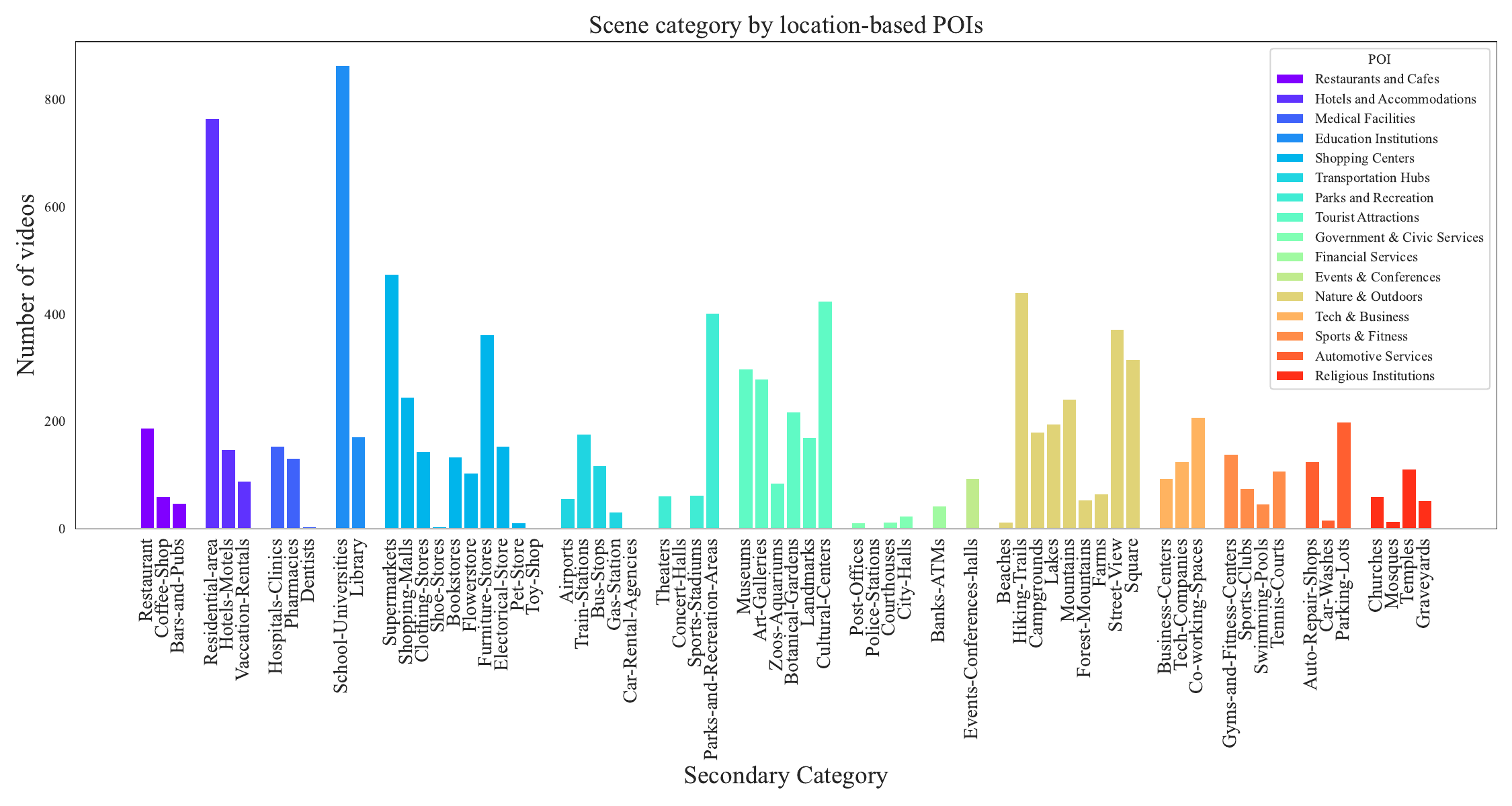}
    \caption{Number of scenes within secondary POI category. The legend contains the mapping between the primary and secondary POI categories. We observe that '\textit{schools-universities} and '\textit{residential-area}' are the predominant scenes in our \name \ dataset. In contrast, locations such as government and civic service facilities (e.g., '\textit{post office}, '\textit{police station}, '\textit{court house}, and '\textit{city hall}) are less frequently captured due to the challenges in accessing these areas for detailed video recording. }
    \label{fig:secondary_POI_category}
\end{figure*}

\begin{figure*}[t]
    \centering
    \includegraphics[width=\linewidth]{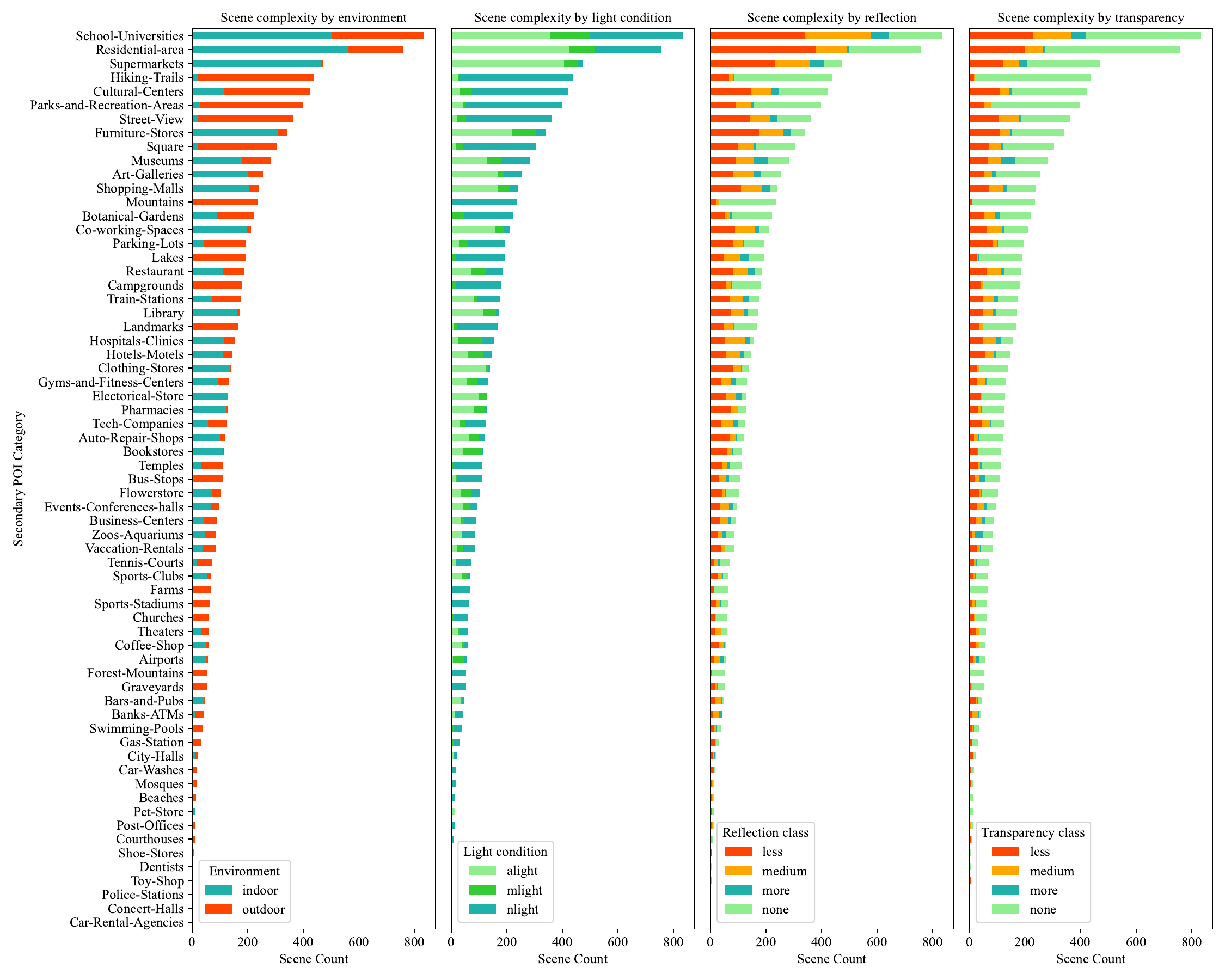}
    \caption{We show the distribution of scenes captured in secondary POI categories by complexities, including environmental setting, light conditions, reflective surfaces, and transparent materials. }
    \label{fig:secondary_POI_compleixty}
\end{figure*}

\paragraph{Scene summary by secondary POI category.} The secondary POI categories are detailed classes within the primary POI categories. Fig.~\ref{fig:secondary_POI_category} shows scene statistics for each secondary POI category and the corresponding primary POI category.  Fig.~\ref{fig:secondary_POI_compleixty} presents scene statistics for each secondary POI category by complexity indices such as environmental setting, light condition, and level of reflection and transparency. For example, in the '\textit{light condition}' attribute, we find that scenes from '\textit{supermarkets}', '\textit{shopping-malls}, and '\textit{furniture-stores'} are mostly under artificial lighting, whereas \textit{'hiking-trails} and \textit{'parks-and-recreation-areas'} are under natural light. As for '\textit{reflection}' and '\textit{transparency}' attributes, '\textit{shopping-malls'} are more likely to feature fully reflective scenes than other locations, while nature \& outdoor scenes such as '\textit{hiking-trails}' are predominantly non-reflective scenes. Most scenes are non-transparent. These observations align well with common expectations in real-world scenarios. We present a sample of \name\ in the \textbf{video demo}.

\begin{figure*}[t]
    \centering
    \includegraphics[width=\linewidth]{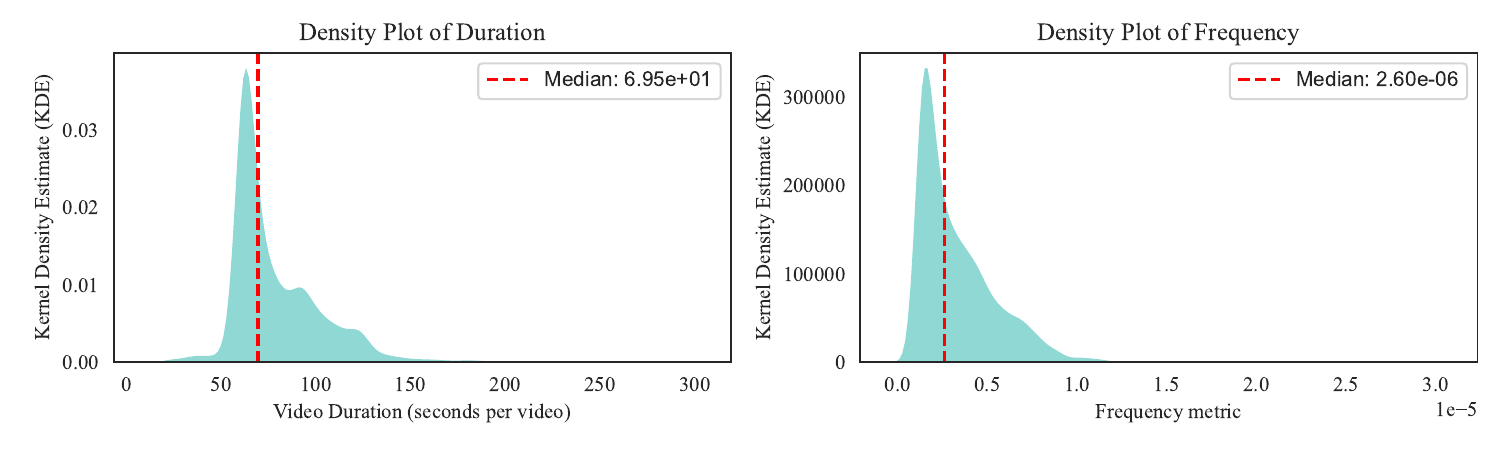}
    \caption{We show the distribution of video duration and frequency metric in \scale\ videos. 
    The minimum duration for video shooting with consumer mobile devices is set at 60 secs, while for drone cameras, it's at least 45 secs. In our dataset, the median video duration is 69.5 secs. Furthermore, the median value of the frequency metric, determined by the average image intensity, stands at 2.6e-06. Based on this median value, we categorize scenes into high frequency ('\textit{high\_freq}') and low frequency ('\textit{low\_freq}') classes.}
    \label{fig:dur_freq_distribution}
\end{figure*}

\paragraph{Frequency and duration estimates.} The kernel density distribution of frequency metric and video duration can be found in Fig.~\ref{fig:dur_freq_distribution}. The frequency classes are delineated based on the median value of the frequency metric.

\section{Experiment}\label{sec.exp}
\subsection{NVS benchmark}
\paragraph{Experiment Details}
The implementation of Nerfacto and Instant-NGP is from nerfstudio~\cite{tancik2023nerfstudio}. 
MipNeRF360~\cite{barron2022mip} and 3D gaussian splatting (3DGS)~\cite{kerbl20233d} codes are from the authors. 
ZipNeRF~\cite{barron2023zip} source code is not public yet when we submit the paper. 
We used a public implementation~\cite{zipnerf-pytorch} that shows the same performance results reported in the paper to test ZipNeRF.

The default ray batch is 4096. ZipNeRF is sensitive to this parameter, and we also showed 65536 (default by ZipNeRF) results. Nerfacto, Instant-NGP, ZipNeRF used half-precision fp16 while 3DGS and MipNeRF360 use full precision. 
All the NeRF-based methods use the same near ($0.01$) and the same far ($1e5$).
The codes are run on A30, V100 and A100 GPUs depending on the memory they used. 
All the experiments took about 13,230 GPU hrs to finish.

\begin{figure*}[t]
    \centering
\includegraphics[width=\textwidth]{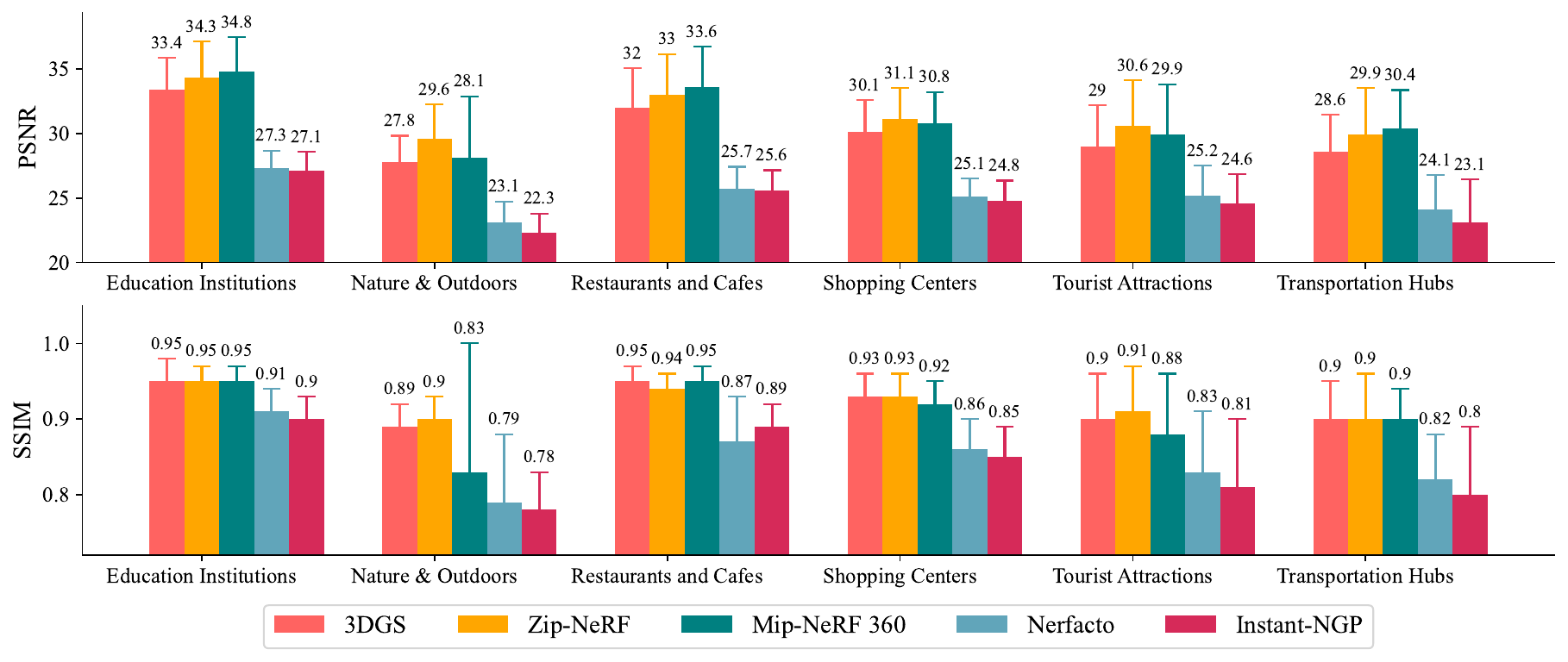}
    \caption{We present the average performance on 6 primary POI categories (\textit{Education institutions, Nature \& Outdoors, Restaurants and Cafes, Shopping Centers, Tourist Attractions, and Transportation Hubs}) in the \benchmark. The text above the bar plot is the mean value of the methods on the primary POI categories. As shown in the figure, NVS methods have better performance on scenes captured in \textit{Education institutions}, \textit{Restaurants and Cafes}, \textit{Shopping Centers} than \textit{Tourist Attractions}, \textit{Transportation Hubs}, and \textit{Nature \& Outdoors}. Because majority scenes in \textit{Education institutions}, \textit{Restaurants and Cafes}, and \textit{Shopping Centers} are indoor scenes. Additionally, the performance on \textit{Shopping Centers} is worse than textit{Education institutions} and \textit{Restaurants and Cafes}.  }
    \label{fig:overall_method_scene_diversity}
\end{figure*}

\paragraph{More quantitative results.} We present the performance of State-of-the-art (SOTAs) in \benchmark\ by scene  primary POI categories in Fig.~\ref{fig:overall_method_scene_diversity}. 

\begin{figure*}[t]
    \centering
    \includegraphics[trim=140 0 0 50, clip, width=\linewidth]{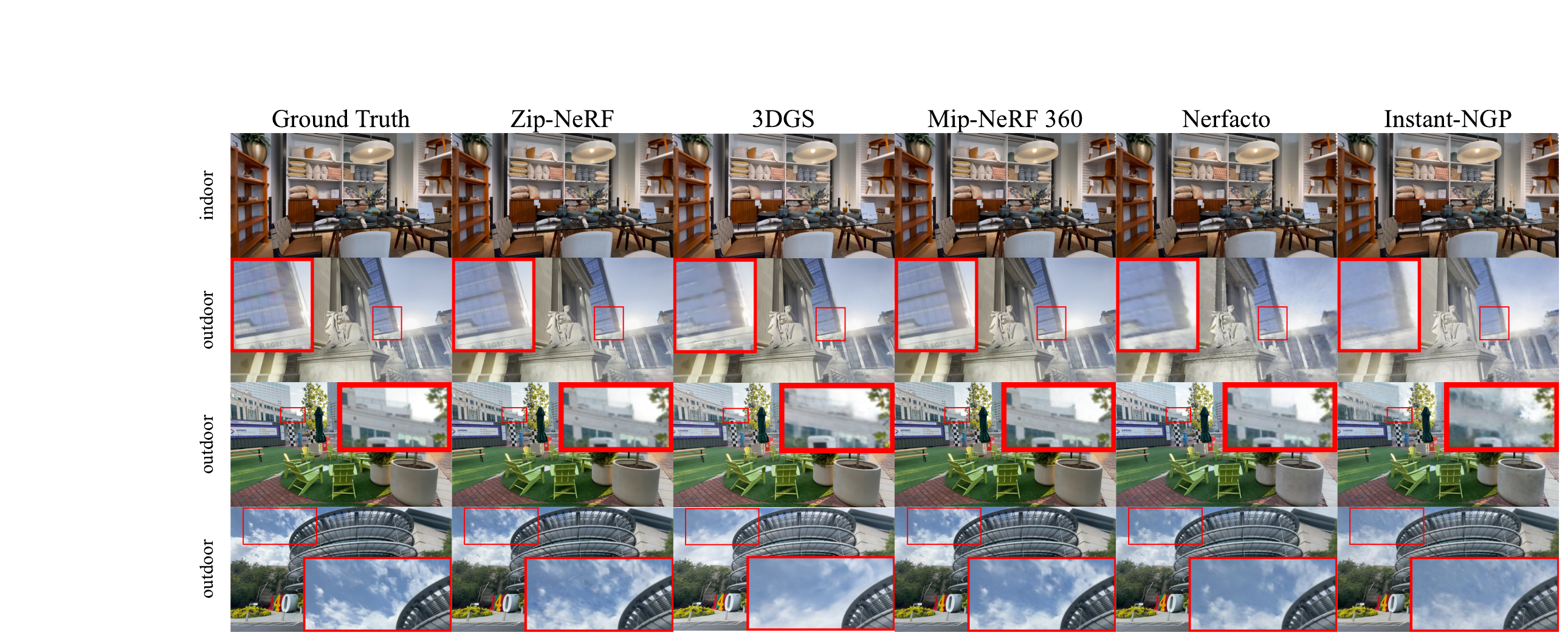}
    \caption{We compare the SOTAs for indoor (bounded) and outdoor (unbounded) environments on \benchmark\ from held-out test views. As illustrated in the figure, indoor scenes pose fewer challenges compared to outdoor scenes, where SOTAs demonstrate varying levels of performance. We observe that outdoor scene is more challenging for 3D Gaussian Splatting (3DGS), Nerfacto, and Instant-NGP than Zip-NeRF and Mip-NeRF 360.}
    \label{fig:supp_indoor_outdoor}
\end{figure*}

\begin{figure*}[t]
    \centering
    \includegraphics[width=\linewidth]{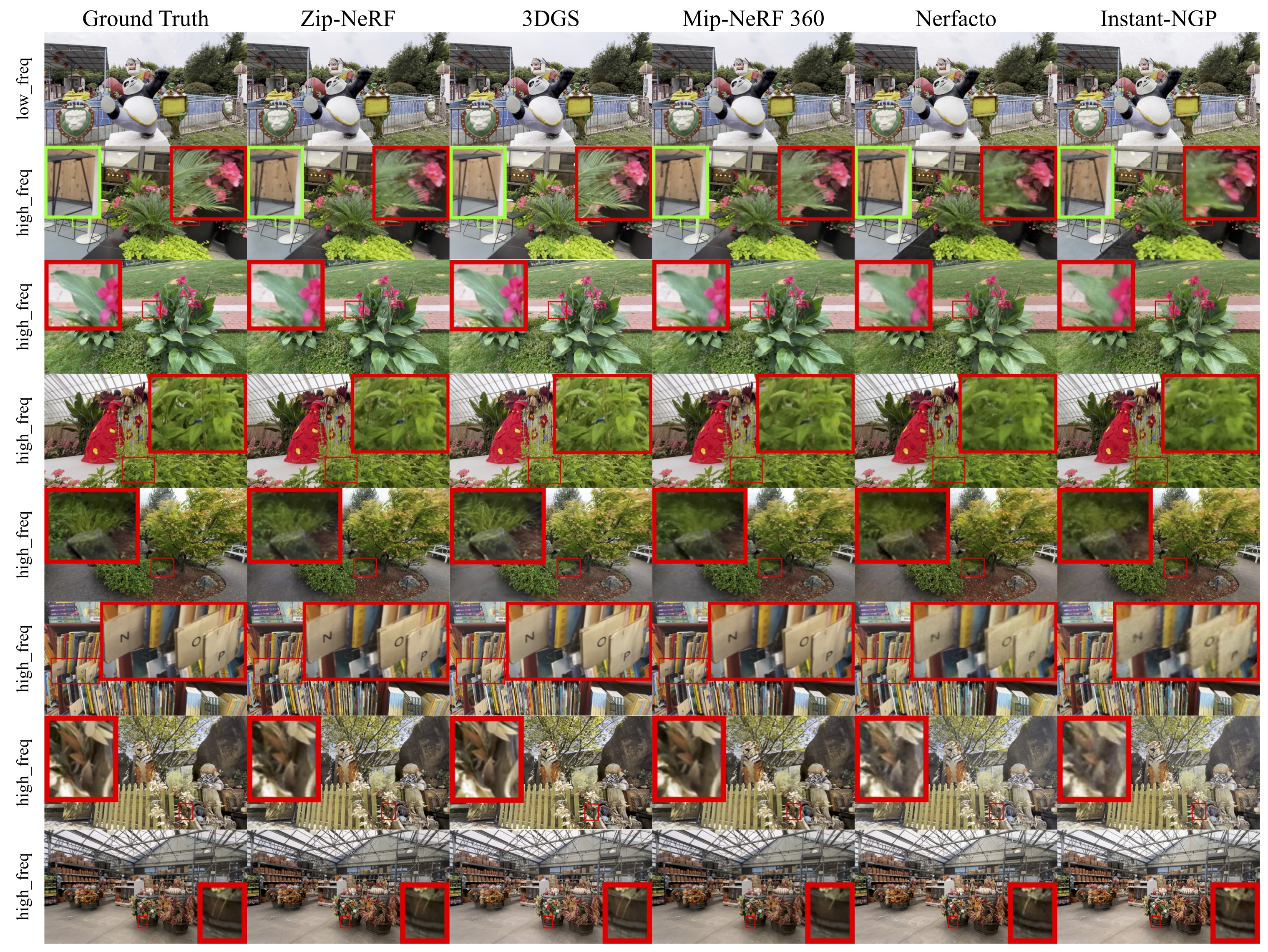}
    \caption{We compare the performance of SOTAs in frequency (\textit{$low\_freq$} vs. \textit{$high\_freq$}) on \benchmark\ from held-out test views. As shown in the figure, high frequency (\textit{high\_freq}) scene is more challenging than low frequency (\textit{low\_freq}) scene. We observe that 3DGS consistently captures scenes with high-frequent details and renders the shape edge for the scene details. As for NeRF variants, it is more challenging for Nerfacto and Instant-NGP to handle scenes with high-frequent details than Zip-NeRF and Mip-NeRF 360. Besides, NeRF variants suffer aliasing issues. }
    \label{fig:supp_frequency}
\end{figure*}

\begin{figure*}[t]
    \centering
    \includegraphics[trim=140 0 0 50, clip, width=\linewidth]{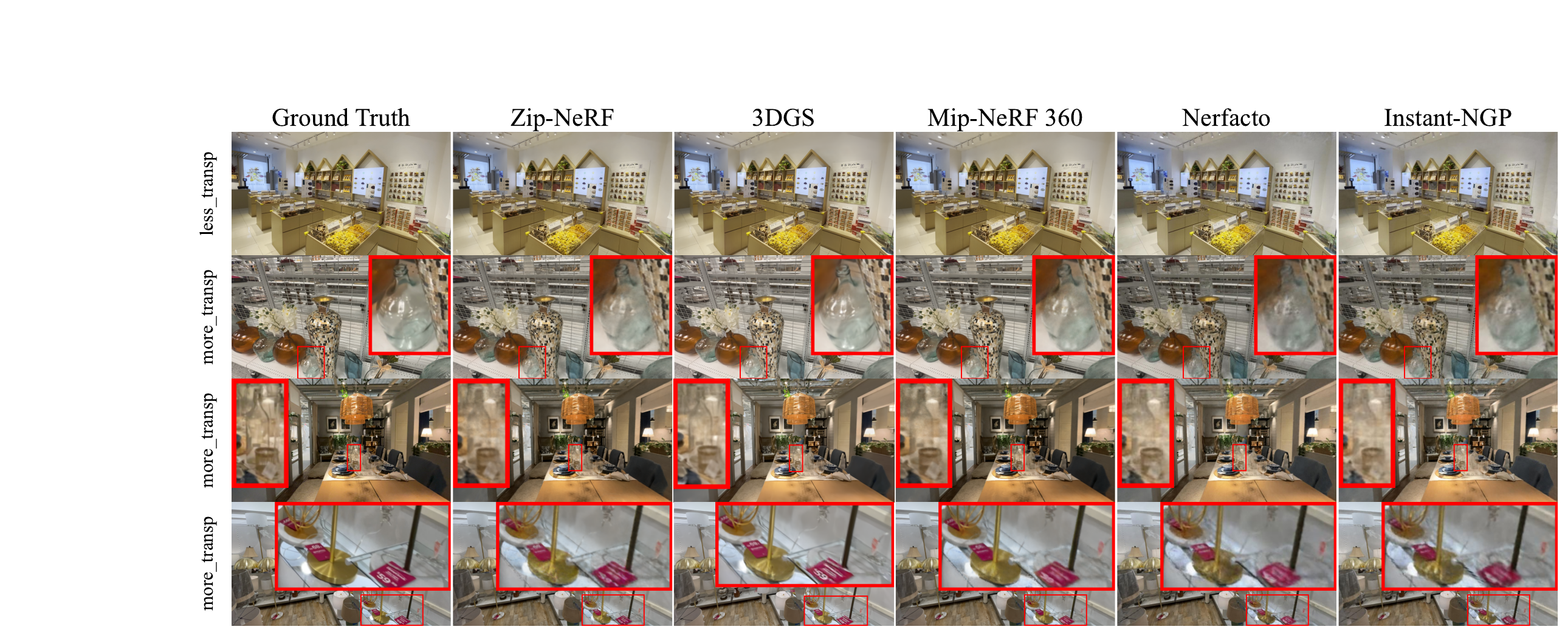}
    \caption{We compare the performance of SOTAs for transparency classes (\textit{$less\_transp$} vs. \textit{$more\_transp$}) on \benchmark\ from held-out test views. As shown in the figure, scenes with more transparent materials (\textit{more\_transp}) are more challenging than scenes with less transparent materials (\textit{less\_transp}). In our analysis of the selected scenes, we noted that 3DGS, Zip-NeRF, and Mip-NeRF 360 effectively capture the subtle edges of transparent objects. Conversely, Nerfacto and Instant-NGP tend to consistently generate artifacts.}
    \label{fig:supp_transparency}
\end{figure*}

\begin{figure*}[t]
    \centering
    \includegraphics[trim=140 0 0 50, clip, width=\linewidth]{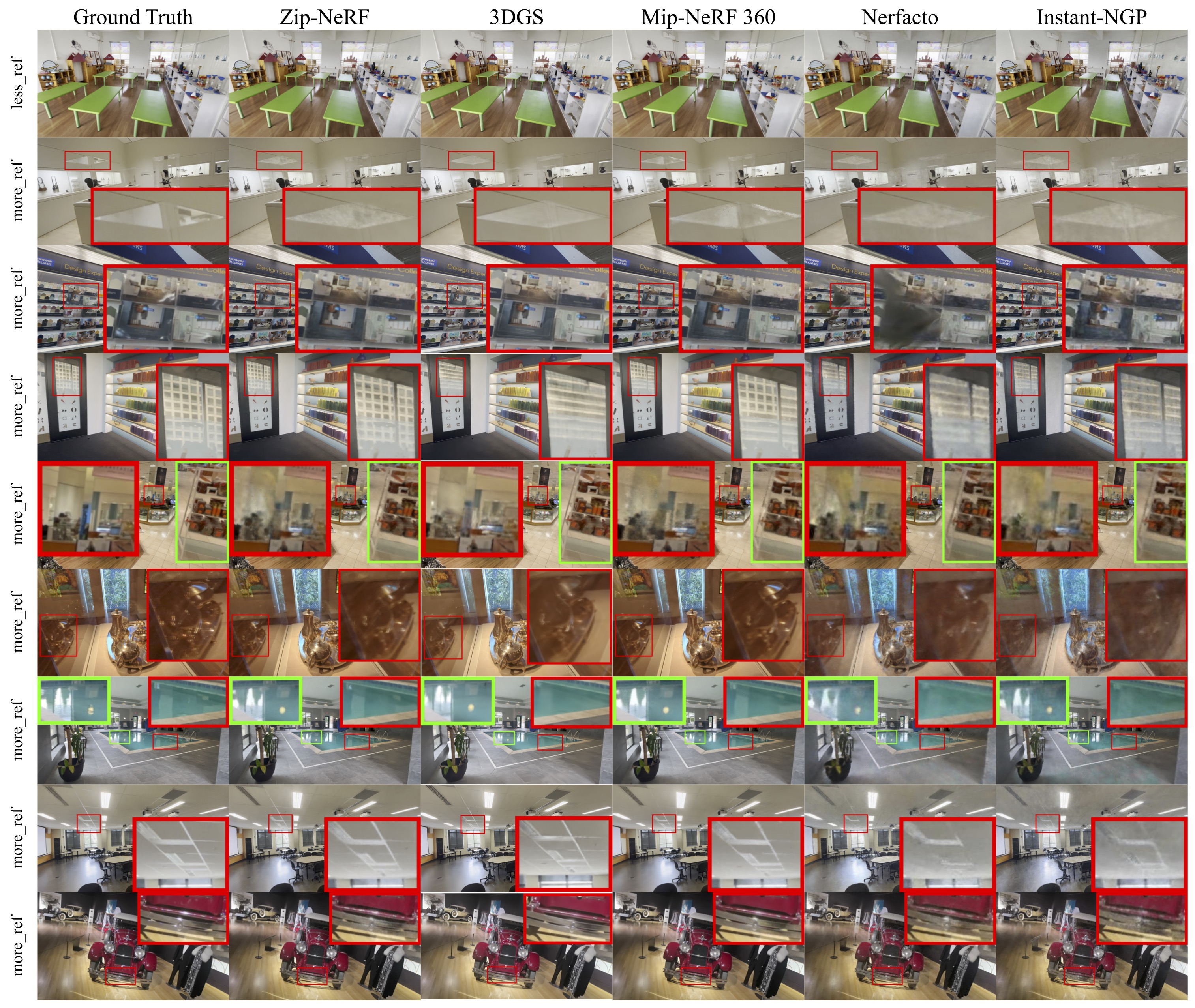}
    \caption{We compare the SOTAs for reflection classes (\textit{$less\_ref$} vs. \textit{$more\_ref$}) on \benchmark\ from held-out test views. As shown in the figure, scenes with more reflective surfaces (\textit{more\_ref}) are more challenging than scenes with less reflective surfaces (\textit{less\_ref}). Among SOTAs, Zip-NeRF and Mip-NeRF 360 are adept at capturing subtle reflections and highlights. On the other hand, 3DGS tends to overly smooth out less intense reflections. Nerfacto and Instant-NGP struggle to effectively manage scenes with highly reflective surfaces, often resulting in the generation of artifacts.}
    \label{fig:supp_reflection}
\end{figure*}

\paragraph{More visual results.} We present more visual results for the performance of SOTAs on \benchmark\ by scene complexity indices. In particular, Fig.~\ref{fig:supp_indoor_outdoor} describes the performance of SOTAs by environmental setting; Fig.~\ref{fig:supp_frequency} describes the performance of SOTAs by frequency; Fig.~\ref{fig:supp_transparency} describes the performance of SOTAs by transparency; and Fig.~\ref{fig:supp_reflection} describes the performance of SOTAs by reflection. 
Due to the file size limitation, we provide visual results for 70 scenes in \benchmark\ in the \textbf{HTML} supplementary submission.

\subsection{Generalizable NeRF}
\paragraph{Experiment details}
We follow the default setting by IBRNet~\cite{wang2021ibrnet}. 
The training dataset includes LLFF~\cite{mildenhall2019local}, spaces, RealEstate10K~\cite{zhou2018stereo} and self-collected small dataset by IBRNet authors. 
The evaluation dataset includes Diffuse Synthetic $360^\circ$~\cite{sitzmann2019scene}, Realistic Synthetic $360^\circ$~\cite{mildenhall2021nerf}, part of LLFF that was not used in training.
We used the official implementation.
Each experiment was trained on single A100 GPU. 
Pretaining on Scannet++ and \name\ took 24 hrs. 

\begin{figure*}[t]
    \setlength{\tabcolsep}{1pt} %
    \centering
    \begin{tabular}{cccc}
    \includegraphics[width=0.24\linewidth]{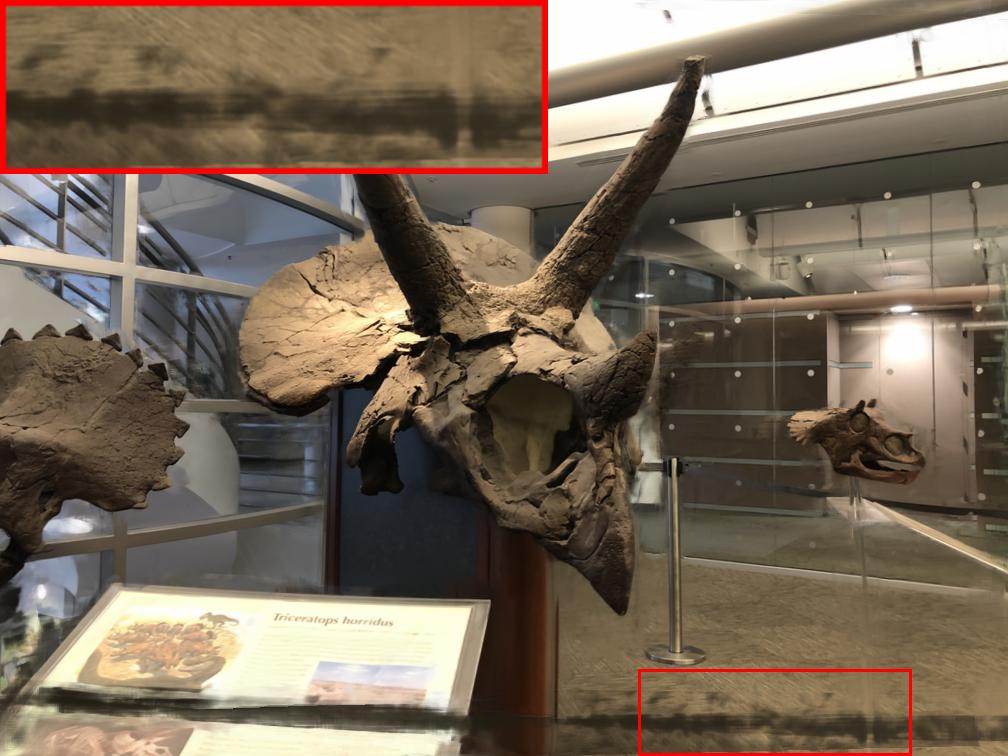} &
    \includegraphics[width=0.24\linewidth]{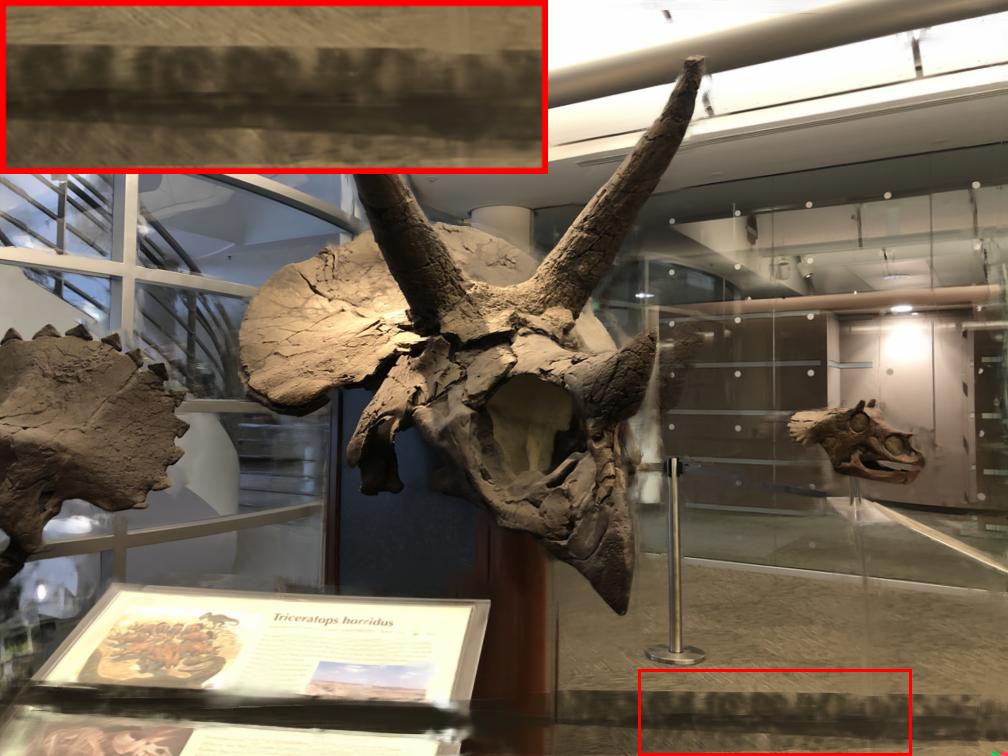} &
    \includegraphics[width=0.24\linewidth]{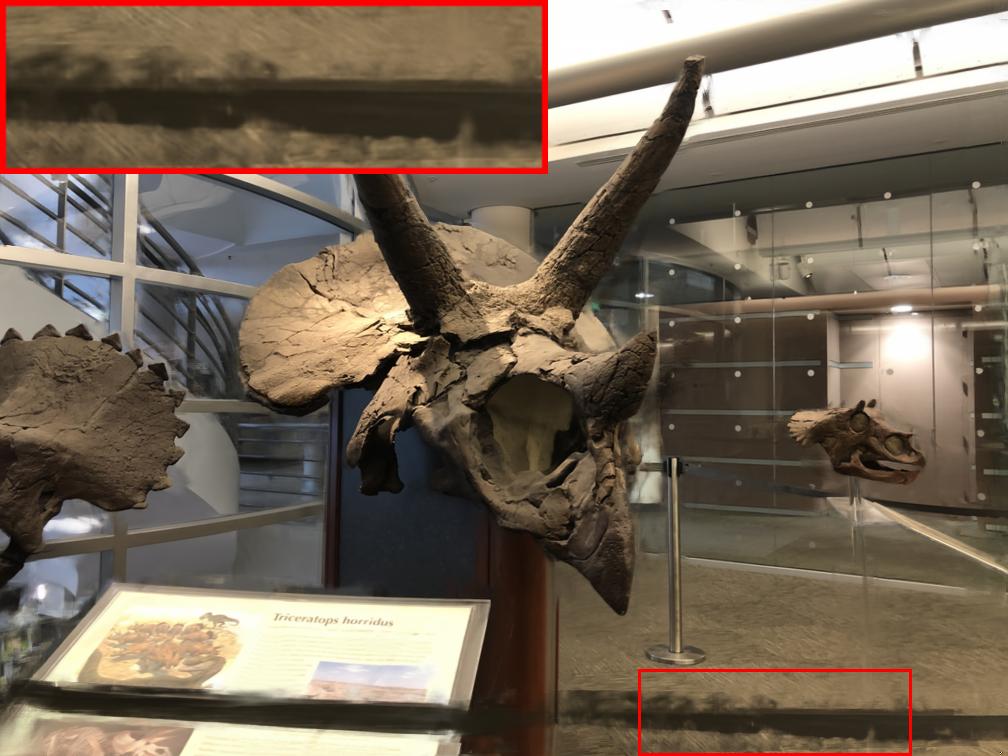} &
    \includegraphics[width=0.24\linewidth]{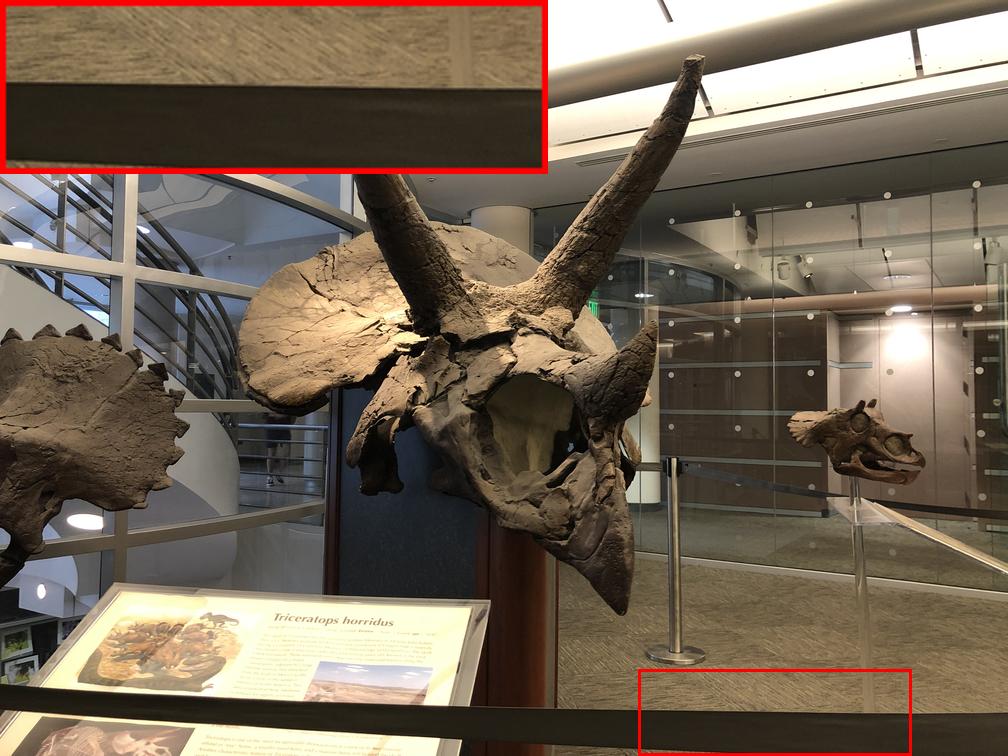} \\
    \includegraphics[width=0.24\linewidth]{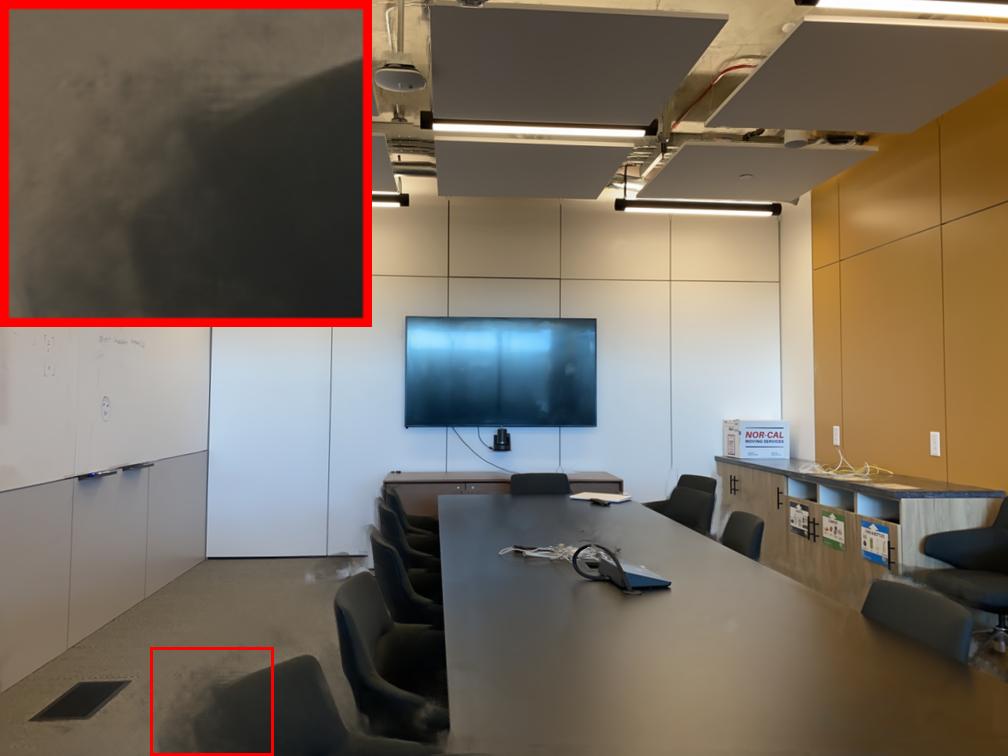} &
    \includegraphics[width=0.24\linewidth]{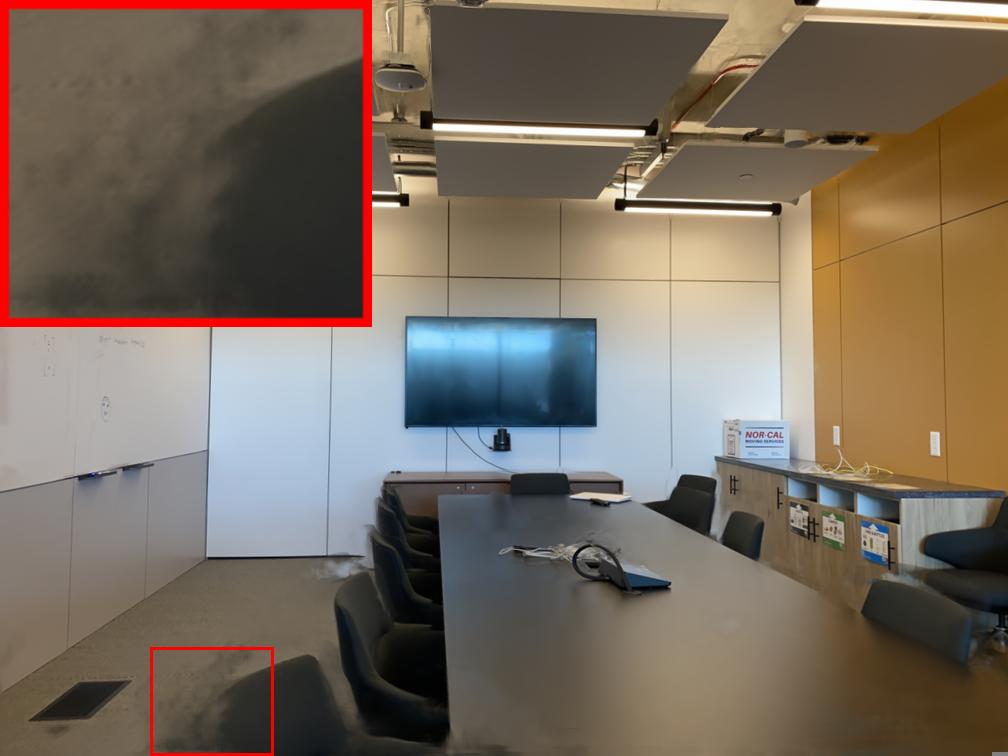} &
    \includegraphics[width=0.24\linewidth]{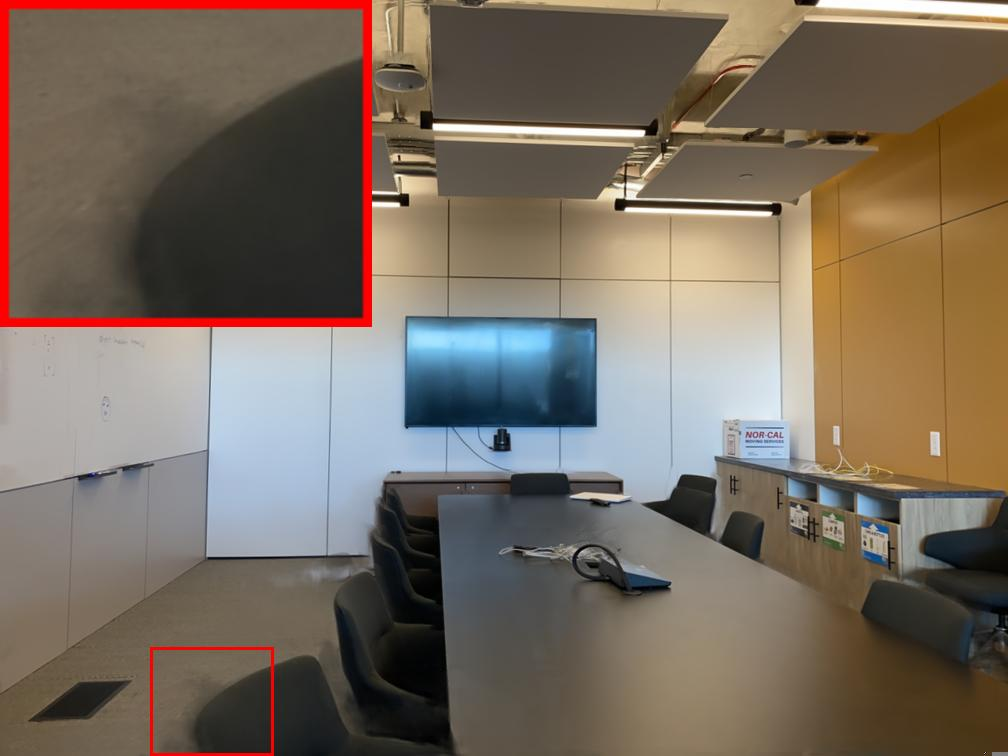} & 
    \includegraphics[width=0.24\linewidth]{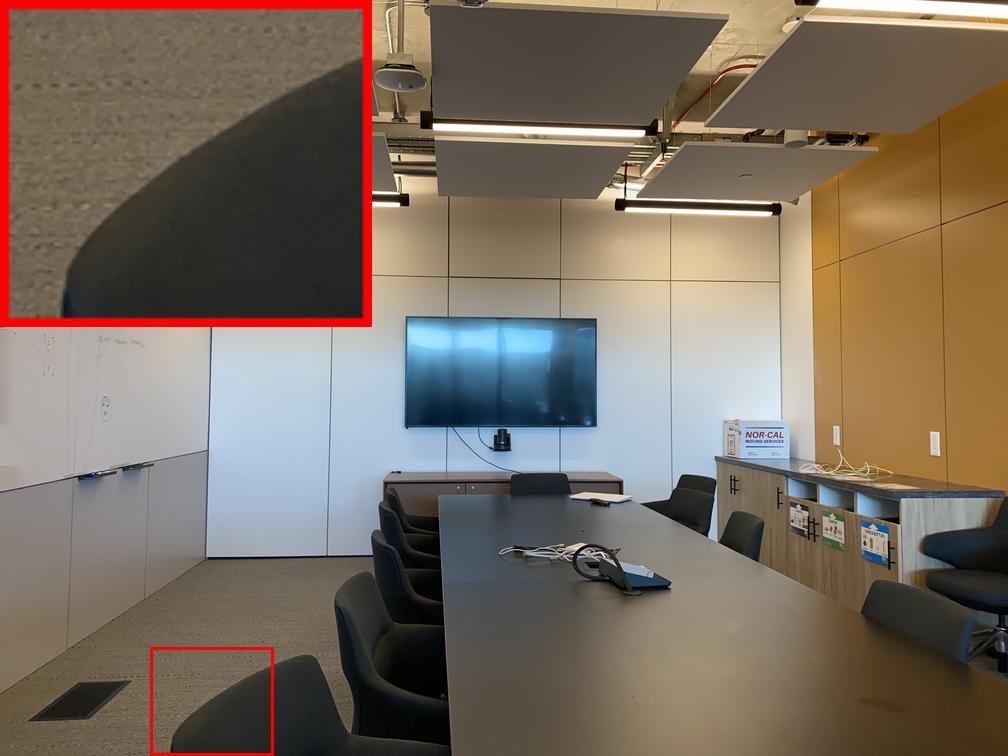} \\
    IBRNet & IBRNet-S & IBRNet-D & GT \\
    \end{tabular}
    \caption{\textbf{More qualitative results for generalizable NeRF.} IBRNet-S: pretrain IBRNet on Scannet++~\cite{yeshwanth2023scannet++}. IBRNet-D: pretrain IBRNet on \name{}. Priors learned from \name{} help IBRNet perform the best on the evaluation.}
    \label{fig:enter-label}
\end{figure*}

